\documentclass[preprint]{elsarticle}
\usepackage{geometry} 

\usepackage{amsmath,amsfonts,amssymb}
\usepackage[dvipsnames]{xcolor}
\usepackage{bm,relsize}
\usepackage{physics}
\usepackage{graphicx}
\usepackage{blindtext}
\usepackage{pgfplots}
\usepackage{array}
\usepackage{graphicx}
\usepackage{multirow}
\usepackage{algpseudocode,algorithm}
\usepackage[hidelinks]{hyperref}
\usepackage[capitalize]{cleveref}
\crefname{equation}{}{}
\numberwithin{equation}{section}
\crefformat{appendix}{#2#1#3}

\newcommand{\NN}{{\mathsmaller{\mathcal N}}}
\renewcommand{\bf}[1]{\mathbf{#1}}
\newcommand{\ol}[1]{\overline{#1}}
\newcommand{\wt}[1]{\widetilde{#1}}
\newcommand{\filter}{\mathcal F_{\bar{\mathsmaller{\Delta}}}}
\newcommand{\filtersize}{{\bar{\mathsmaller{\Delta}}}}
\newcommand{\scf}{\textsc{f}}

\definecolor{C1}{HTML}{D53288}
\definecolor{C2}{HTML}{DC8045}
\definecolor{C3}{HTML}{21B14B}
\definecolor{C4}{HTML}{008AC2}
\definecolor{C5}{HTML}{3F459B}

\colorlet{diff}{RedOrange!90!white}
\colorlet{anti}{BlueGreen!90!black}
\colorlet{zero}{Lavender}

\pgfplotsset{
	compat=1.17,
	tvd/.style={
		C4,semithick
	},
	no/.style={
		C1,semithick
	},
	init/.style={
		gray,semithick
	},
	exact/.style={
		C3,semithick
	},
	tvdmark/.style={
		C4,thick,mark=*,mark options={fill=white,scale=1.5}
	},
	nomark/.style={
		C1,thick,mark=x,mark options={scale=1.5}
	},
	initmark/.style={
		gray,thick,mark=diamond*,mark options={fill=white,scale=1.5}
	},
	exactmark/.style={
		C3,thick,mark=square*,mark options={fill=white,scale=1.5}
	}
}
\usetikzlibrary{spy}
\pgfkeys{/pgf/number format/.cd,1000 sep={}}
\RequirePackage[normalem]{ulem} 
\RequirePackage{color}\definecolor{RED}{rgb}{1,0,0}\definecolor{BLUE}{rgb}{0,0,1} 
\providecommand{\DIFaddbegin}{} 
\providecommand{\DIFaddend}{} 
\providecommand{\DIFdelbegin}{} 
\providecommand{\DIFdelend}{} 
\providecommand{\DIFaddbeginFL}{} 
\providecommand{\DIFaddendFL}{} 
\providecommand{\DIFdelbeginFL}{} 
\providecommand{\DIFdelendFL}{} 
\newcommand{\DIFscaledelfig}{0.5}
\RequirePackage{settobox} 
\RequirePackage{letltxmacro} 
\newsavebox{\DIFdelgraphicsbox} 
\newlength{\DIFdelgraphicswidth} 
\newlength{\DIFdelgraphicsheight} 
\LetLtxMacro{\DIFOincludegraphics}{\includegraphics} 
\newcommand{\DIFaddincludegraphics}[2][]{{\color{blue}\fbox{\DIFOincludegraphics[#1]{#2}}}} 
\newcommand{\DIFdelincludegraphics}[2][]{
\sbox{\DIFdelgraphicsbox}{\DIFOincludegraphics[#1]{#2}}
\settoboxwidth{\DIFdelgraphicswidth}{\DIFdelgraphicsbox} 
\settoboxtotalheight{\DIFdelgraphicsheight}{\DIFdelgraphicsbox} 
\scalebox{\DIFscaledelfig}{
\parbox[b]{\DIFdelgraphicswidth}{\usebox{\DIFdelgraphicsbox}\\[-\baselineskip] \rule{\DIFdelgraphicswidth}{0em}}\llap{\resizebox{\DIFdelgraphicswidth}{\DIFdelgraphicsheight}{
\setlength{\unitlength}{\DIFdelgraphicswidth}
\begin{picture}(1,1)
\thicklines\linethickness{2pt} 
{\color[rgb]{1,0,0}\put(0,0){\framebox(1,1){}}}
{\color[rgb]{1,0,0}\put(0,0){\line( 1,1){1}}}
{\color[rgb]{1,0,0}\put(0,1){\line(1,-1){1}}}
\end{picture}
}\hspace*{3pt}}} 
} 
\LetLtxMacro{\DIFOaddbegin}{\DIFaddbegin} 
\LetLtxMacro{\DIFOaddend}{\DIFaddend} 
\LetLtxMacro{\DIFOdelbegin}{\DIFdelbegin} 
\LetLtxMacro{\DIFOdelend}{\DIFdelend} 
\DeclareRobustCommand{\DIFaddbegin}{\DIFOaddbegin \let\includegraphics\DIFaddincludegraphics} 
\DeclareRobustCommand{\DIFaddend}{\DIFOaddend \let\includegraphics\DIFOincludegraphics} 
\DeclareRobustCommand{\DIFdelbegin}{\DIFOdelbegin \let\includegraphics\DIFdelincludegraphics} 
\DeclareRobustCommand{\DIFdelend}{\DIFOaddend \let\includegraphics\DIFOincludegraphics} 
\LetLtxMacro{\DIFOaddbeginFL}{\DIFaddbeginFL} 
\LetLtxMacro{\DIFOaddendFL}{\DIFaddendFL} 
\LetLtxMacro{\DIFOdelbeginFL}{\DIFdelbeginFL} 
\LetLtxMacro{\DIFOdelendFL}{\DIFdelendFL} 
\DeclareRobustCommand{\DIFaddbeginFL}{\DIFOaddbeginFL \let\includegraphics\DIFaddincludegraphics} 
\DeclareRobustCommand{\DIFaddendFL}{\DIFOaddendFL \let\includegraphics\DIFOincludegraphics} 
\DeclareRobustCommand{\DIFdelbeginFL}{\DIFOdelbeginFL \let\includegraphics\DIFdelincludegraphics} 
\DeclareRobustCommand{\DIFdelendFL}{\DIFOaddendFL \let\includegraphics\DIFOincludegraphics} 
\RequirePackage{listings} 
\RequirePackage{color} 
\lstdefinelanguage{DIFcode}{ 
  moredelim=[il][\color{red}\sout]{\%DIF\ <\ }, 
  moredelim=[il][\color{blue}\uwave]{\%DIF\ >\ } 
} 
\lstdefinestyle{DIFverbatimstyle}{ 
	language=DIFcode, 
	basicstyle=\ttfamily, 
	columns=fullflexible, 
	keepspaces=true 
} 
\lstnewenvironment{DIFverbatim}{\lstset{style=DIFverbatimstyle}}{} 
\lstnewenvironment{DIFverbatim*}{\lstset{style=DIFverbatimstyle,showspaces=true}}{} 

\begin{document}
\begin{frontmatter}
\title{A TVD neural network closure and application to turbulent combustion}
\author[1]{Seung Won Suh\corref{cor1}}
\ead{suh29@illinois.edu}
\author[2]{Jonathan F. MacArt}
\ead{jmacart@nd.edu}
\author[3]{Luke N. Olson}
\ead{lukeo@illinois.edu}
\author[4]{Jonathan B. Freund}
\ead{jbfreund@illinois.edu}
\cortext[cor1]{Corresponding author}
\affiliation[1]{organization={Mechanical Science and Engineering, University of Illinois Urbana--Champaign},
	addressline={1205 W. Clark St.},
	postcode={61801},
	city={Urbana, Illinois},
	country={United States of America}}
\affiliation[2]{organization={Aerospace and Mechanical Engineering, University of Notre Dame},
	addressline={365 Fitzpatrick Hall of Engineering},
	postcode={46556},
	city={Notre Dame, Indiana},
	country={United States of America}}
\affiliation[3]{organization={Computer Science, University of Illinois Urbana--Champaign},
	addressline={201 N. Goodwin Ave.},
	postcode={61801},
	city={Urbana, Illinois},
	country={United States of America}}
\affiliation[4]{organization={Aerospace Engineering, University of Illinois Urbana--Champaign},
	addressline={104 S. Wright St.},
	postcode={61801},
	city={Urbana, Illinois},
	country={United States of America}}
\begin{abstract}
Trained neural networks (NN) have attractive features for closing governing equations.
There are many methods that are showing promise, but all can fail in cases when small errors consequentially violate physical reality, such as a solution boundedness condition.
A NN formulation is introduced to preclude spurious oscillations that violate solution boundedness or positivity.
It is embedded in the discretized equations as a machine learning closure and strictly constrained, inspired by total variation diminishing (TVD) methods for hyperbolic conservation laws.
The constraint is exactly enforced during gradient-descent training by rescaling the NN parameters, which maps them onto an explicit feasible set.
Demonstrations show that the constrained NN closure model usefully recovers linear and nonlinear hyperbolic phenomena and anti-diffusion while enforcing the non-oscillatory property.
Finally, the model is applied to subgrid-scale (SGS) modeling of a turbulent reacting flow, for which it suppresses spurious oscillations in scalar fields that otherwise violate the solution boundedness.
It outperforms a simple penalization of oscillations in the loss function.
\end{abstract}
\end{frontmatter}
\section{Introduction} \label{sec:intro}

Fully-resolved simulations of multiscale physical systems are often infeasible due to computational cost.
To alleviate this, small scales are often unresolved or unrepresented, which reduces the predictive accuracy of the simulation.
Many partial differential equation (PDE) models also employ reduced or uncertain models; for example,
PDE models for reacting flow typically use simplified models for species diffusion and chemical kinetics~\cite{williams2018combustion}.
With these factors degrading the fidelity of a simulation, it is desirable to represent unresolved and unknown or neglected physics using computationally efficient models.

Our starting point is an acceptance that machine learning (ML) is attractive for this PDE closure challenge due to its model-form flexibility and the rapid advance of software and hardware support.
We pursue a class of ML closures that embed a trainable neural network (NN) in a PDE model to represent missing or inaccurate terms~\cite{sirignano2020dpm,um2020solver}.
Specifically, we consider the following type of closure modeling problem:
\begin{subequations}
	\begin{align}
		\text{find} \quad &\vec \theta \\
		\text{that minimizes} \quad &\mathcal J [q, h, \vec \theta\,] \label{eqn:loss-general} \\
		\text{subject to} \quad &\partial_t q = R(q) + h(q; \vec \theta\,), \label{eqn:problem}
	\end{align}
\end{subequations}
where $q(\bf x,t)$ is a time $t$ evolving solution over a space $\bf x$, $R(q)$ is the right-hand-side (RHS) term of the unclosed PDE, and $h(q; \vec \theta\,)$ is a NN closure with $N_\theta$ trainable parameters $\vec \theta \in\mathbb{R}^{N_\theta}$.
Altogether, \eqref{eqn:problem} is a NN-augmented PDE---serving as an optimization constraint---with $\mathcal J [q, h, \vec \theta\,] \in \mathbb{R}$ in~\cref{eqn:loss-general} the loss functional.
This approach has been pursued for ML-based turbulence modeling with both \textit{a priori}~\cite{ling2016reynolds,wu2018physics,park2021toward,ling2016machine} and \textit{a posteriori}~\cite{sirignano2020dpm,um2020solver,macart2021embedded} training.
Note that it is different from other ML strategies for PDEs, such as the Physics-Informed Neural Network (PINN) approach~\cite{raissi2019physics}, the Deep Operator Network (DeepONet)~\cite{lu2021learning}, or the Deep Galerkin Method (DGM)~\cite{sirignano2018dgm}, all of which train NNs to produce the solution directly without solving the PDE.
We take as given that such model $h(q; \vec \theta\,)$ will fit closures well enough, except that it can lead to small but catastrophic qualitative violations of physical reality.

Preservation of such solution properties are often required by PDE-embedded models (NN-based or not) for their numerical representation.
For example, local conservation may be preserved exactly by casting the NN-augmented PDE into a discretely conservative form~\cite{macart2021embedded}.
For the case of \eqref{eqn:problem}, conservation may be ensured by replacing $h$ with $-\nabla \cdot \bf f$, where $\nabla$ is the divergence over $\bf x$, and $\bf f$ is a vector-valued NN.
Likewise, invariance properties may be leveraged by selecting invariant model inputs~\cite{ling2016machine}, and the symmetry of a NN model (e.g., to close the Reynolds stress tensor) may be enforced by modeling only independent components~\cite{sirignano2020dpm}.

A more challenging case is when small errors can render a solution qualitatively incorrect.
For any system with sharp features, such as shocks or flames, spurious (numerical) oscillations can violate physical boundedness and positivity constraints.
Preventing this failure mode within a NN closure approach is challenging.
Employing otherwise unneeded mesh resolution to suppress it can be prohibitively expensive, if it can work at all~\cite{ferziger2019computational}.
Numerical solutions of NN-embedded PDEs as in~\cref{eqn:problem} are also susceptible to constraint-violating spurious oscillations, and avoiding them for whatever NN might be employed is the goal of this paper.
We focus primarily on this goal, assuming that the usual concerns about model fidelity for extrapolation is addressed independently by the many schemes that have been proposed~\cite{sirignano2020dpm,um2020solver,ling2016reynolds,wu2018physics,park2021toward,ling2016machine,macart2021embedded}.
Hence, we focus on constraints rather than overall model efficacy.

\begin{figure}
	\centering
	\includegraphics[width=\linewidth]{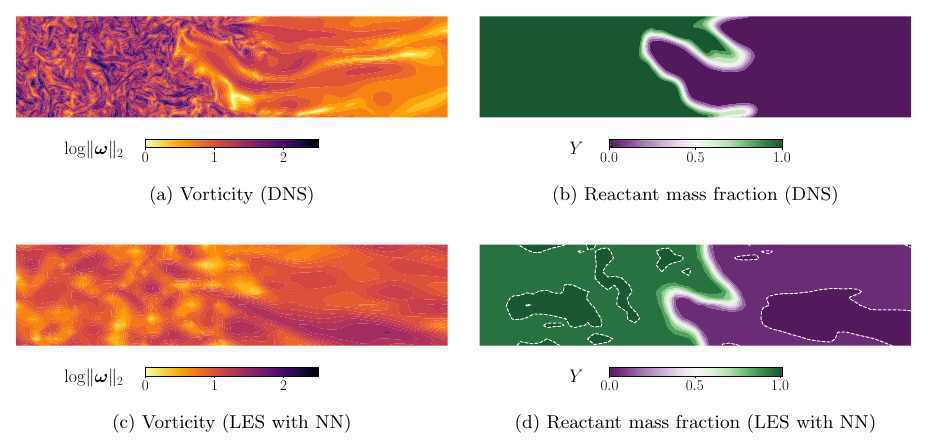}
	\caption{Simulations of three-dimensional turbulent premixed flames: (a, c) vorticity $\bm \omega$ magnitude for (a) DNS, and (c) LES with trained NN closure; and (b, d) mass fraction $Y$ of a reactant species for (b) DNS, and (d) LES with trained NN closure. Dashed contours in (d) indicate $Y=0$ and $Y=1$, and the constraint $Y \in [0,1]$ is violated within the enclosed regions. Contours are plotted on a two-dimensional slice of the three-dimensional domain. The plot (d) is identical to~\cref{fig:flame}(b).}
	\label{fig:example}
\end{figure}

Ideally, NN models could be trained to avoid spurious oscillations without explicitly enforcing the constraint.
For example, if $\mathcal J \equiv D(q,q_e)$ measures the distance between $q$ and a true solution $q_e$, then the model $h$ would maintain smooth $q$ (at least during training) if $\mathcal J = 0$ is exactly achieved.
However, this is generally not the case.
To illustrate this in a complex example, we preview simulation results from~\cref{sec:premixed}.
\Cref{fig:example} visualizes results from a fine-mesh direct numerical simulation (DNS) of turbulent premixed flame and a corresponding coarse-mesh large-eddy simulation (LES) with a trained NN closure model.
The challenge is that the reactant mass fraction $Y$ in~\cref{fig:example}(d) from the LES violates $Y \in [0,1]$, which can lead to additional violations for more complex physics.
Simpler examples in~\cref{sec:unconstrainedNN} and~\cref{sec:demo} will illustrate how unconstrained NN closures suffer from spurious oscillations and how they can be avoided.

To resolve this, the proposed formulation explicitly constrains the model in the optimization procedure:
\begin{subequations}
	\begin{align}
		\text{find} \quad &\vec \theta \\
		\text{that minimizes} \quad &\mathcal J[q, h, \vec \theta\,] \\
		\text{subject to} \quad &\partial_t q = R(q) + h(q; \vec \theta\,) \label{eqn:problem-constrained} \\
		\text{and} \quad &q \in \mathcal Q,
	\end{align}
\end{subequations}
where $\mathcal Q$ is a space of nominally oscillation-free numerical solutions of the NN-augmented PDE~\cref{eqn:problem-constrained}.
Since the smoothness constraint is imposed on the solution $q$, not the NN parameters $\vec \theta$, the constraint cannot be enforced exactly, such as via projected gradient descent method~\cite{levitin1966constrained,bertsekas1997nonlinear}.
There is no means to find a feasible set of $\vec \theta$ that maps all the possible solutions $q$ onto $\mathcal Q$ in order to project $\vec \theta$.
Of course, the constraint violation can be easily penalized via an enriched loss functional $\mathcal J' = \mathcal J + \mathcal P$, where  $\mathcal P \geq 0$ becomes $\mathcal P = 0$ if $q$ satisfies an oscillation-free condition.
However, the resulting model is not guaranteed to exactly satisfy the constraint $\mathcal P = 0$,
because gradient descent generally does not converge to the global optimum for a non-convex loss, and the conflict between $\mathcal J$ and $\mathcal P$ in multi-objective optimization precludes $\mathcal P = 0$~\cite{miettinen1999nonlinear}.

An alternative approach is to design $h$ with constrained parameters $\vec \theta$ to guarantee that $q$ is free of spurious oscillations so $\mathcal P = 0$ exactly.
Such an $h$ maps the $q$-constraint to a $\vec \theta$-constraint as
\begin{equation}
    \vec \theta \in \mathcal C \equiv \left\{ \vec \theta^{\,\prime} \; \vert \; q(\vec \theta^{\,\prime}\,) \in \mathcal Q \right\},
\end{equation}
so it can be strictly enforced on $\vec \theta$ via projected gradient descent:
\begin{equation}
	\vec \theta^{\,k+1} = \pi_{\mathsmaller{\mathcal C}} \left( \vec \theta^{\,k} - \vec \alpha^k \odot \dv{\mathcal J^k}{\vec \theta} \right),
\end{equation}
where $\pi_{\mathsmaller{\mathcal C}}: \mathbb R^{N_\theta} \to \mathcal C$ is a map onto the feasible set $\mathcal C \subseteq \mathbb R^{N_\theta}$, $\vec \alpha$ is the learning rate, and $k$ denotes the gradient-descent iteration.
In enforcing the constraint in \textit{a posteriori} predictions, it is expected to outperform penalization.
At the same time, strict enforcement restricts $h$, though only in the spirit of the universal approximation theorem~\cite{cybenko1989approximation}, which is predicated on a sufficiently effective NN model.
We design and demonstrate a NN model $h$ with parameters $\vec \theta$ that lie in $\mathcal C$, and that is sufficiently flexible to also usefully minimize $\mathcal J$.

Several efforts manifest the benefit of strictly enforcing physical properties on ML models.
Nair \textit{et al.}~\cite{nair2023deep} trained NN closure models for transition-continuum flows with the Clausius--Duhem inequality enforced both strictly and loosely via penalization, showing that the strictly constrained model better extrapolates to out-of-sample flow conditions.
Global hyperbolicity was enforced on ML moment closure models for radiative transfer equations by designing the coefficient matrix to be real diagonalizable~\cite{huang2023machine1} or lower Hessenberg with certain conditions~\cite{huang2023machine2}.
Entropy dissipation and hyperbolicity of the moment system of the Boltzmann equation were ensured by enforcing the closure models to be convex~\cite{schotthofer2022neural,porteous2021data}.
Our approach is in the same spirit as these, though we only focus on the oscillation-free constraint.

The proposed approach complements recent efforts to approximate PDEs with ML.
Chen \textit{et al.}~\cite{chen2024learning} trained a NN flux to approximate one-dimensional hyperbolic PDEs by preserving the solution smoothness via the time integration scheme, but the NN flux closure itself was not limited to exactly preserve the property.
Their previous work~\cite{chen2022deep} made an important step in recognizing the challenge of spurious oscillations in the NN-predicted solutions.
Kim and Kang~\cite{kim2024approximating} demonstrated their Fourier Neural Operator~\cite{li2020fourier} for the same one-dimensional application but without focusing on this challenge.

The design of the new NN model $h$ is discussed in~\cref{sec:method}, where the feasible set $\mathcal C$ is also defined, and it is shown how to constrain $\vec \theta$ onto $\mathcal C$ during optimization.
In~\cref{sec:demo}, the approach is demonstrated on linear and nonlinear advection, a nonlinear hyperbolic system, a non-hyperbolic anti-diffusion problem, and the LES of turbulent reacting flow previewed in~\cref{fig:example}.
Applications in such various scenarios will demonstrate that the method is generalizable.
In~\cref{sec:discussion}, the demonstrated capabilities are discussed along with potential extensions.

\section{Model Design} \label{sec:method}

    \subsection{Unconstrained NN challenge} \label{sec:unconstrainedNN}
		We focus on PDEs for conservation laws, so we use conservation form $h = -\nabla \cdot \bf f$ in~\cref{eqn:problem} in the NN-augmented PDE
		\begin{equation}
			\partial_t q + \nabla \cdot \bf f(q; \vec \theta\,) = R(q)
		\end{equation}
		where $\bf f(q; \vec \theta\,)$ is the trainable NN flux closure.

		We first illustrate the formation of spurious oscillations without the constraint.
		Consider a NN flux that is trained for a solution of a one-dimensional scalar conservation law.
		For simplicity, let $R = 0$, such that the NN model must represent the entirety of the physics.
		The optimization problem is then to
		\begin{subequations} \label{eqn:closure-scalar}
			\begin{align}
				\text{find} \quad &\vec \theta\\
				\text{that minimizes} \quad &\mathcal J = D(q, q_e)\\
				\text{subject to} \quad &\partial_t q + \partial_x f(q; \vec \theta\,) = 0,
				\label{eqn:closure-scalar-pde}
			\end{align}
		\end{subequations}
		where $q(x,t)$ is the solution predicted with the NN flux closure, $q_e(x,t)$ is a trusted (exact if available) solution of the PDE $\partial_t q + \partial_x f_e(q) = 0$ where $f_e$ is the true flux, and $D(q, q_e)$ measures their mismatch.
		Then $\mathcal J \to 0$ implies $q \to q_e$ as well as $f \to f_e$.

		For this demonstration, the model is trained for advection with unit speed $f_e(q) = q$ in a periodic domain $x \in [0,1]$ for $t \in [0, t_f]$, where $t_f$ is the final simulation time.
		Simple finite differencing of~\cref{eqn:closure-scalar-pde} yields
		\begin{equation} \label{eqn:discrete-scalar}
			q_i^n = q_i^{n-1} - \frac{\Delta t}{\Delta x} \left( f_{i+\frac{1}{2}}^{n-1} - f_{i-\frac{1}{2}}^{n-1} \right)
		\end{equation}
		for $i \in \{ 0, 1, \ldots, N_x-1 \}$ and $n \in \{ 1, \ldots, N_t \}$, starting from the initial condition $q_i^0$.
		Space is discretized as $x_i = i \Delta x$ with $\Delta x = 1/N_x$ where $N_x$ is the number of independent grid points, and time is discretized as $t^n = n \Delta t$ with $\Delta t = t_f/N_t$ where $N_t$ is the number of time steps.
		We use $N_x = 100$ so $\Delta x = 10^{-2}$, and $\Delta t = 2.5 \times 10^{-3}$.
		The discrete solution $q_i^n$ approximates $q(x_i,t^n)$, and $f_{i + \frac{1}{2}}^n$ is the NN flux evaluated at the cell face $x_{i+\frac{1}{2}}$ using a two-point stencil:
		\begin{equation} \label{eqn:nn-unconstrained}
			f_{i+\frac{1}{2}}^n = f_\NN(q^n_i, q^n_{i+1}; \vec \theta\,).
		\end{equation}
		The flux $f_\NN$ is a NN with a fully-connected feed-forward architecture~\cite{sirignano2020dpm}:
		\begin{equation} \label{eqn:nn-architecture}
			\begin{split}
				z^1 &= \tanh(W^0 y + b^0) \\
				z^2 &= \tanh(W^1 z^1 + b^1) \\
				z^3 &= z^2 \odot g^1 \quad \text{with} \quad g^1 = \tanh(W^2 y + b^2) \\
				z^4 &= \tanh(W^3 z^3 + b^3) \\
				z^5 &= z^4 \odot g^2 \quad \text{with} \quad g^2 = \tanh(W^4 y + b^4) \\
				\mathrm{NN}(y; \vec \theta\,) &= W^5 z^5 + b^5
			\end{split}
		\end{equation}
		with parameters $\vec \theta \equiv \{W^i; b^i\}_{i=0}^5$.
		The two element-wise products $\odot$ make the model capable of exactly representing third-order polynomials, while the tanh activation function will generally introduce additional nonlinearity~\cite{sirignano2018dgm}.
		We define a space of NN functions
		\begin{equation}
			\begin{aligned}
				\mathcal N (N_{\text{i}} \to N_{\text{h}} \to N_{\text{o}}) \equiv \{ \mathrm{NN}(y;\vec \theta\,) \; \vert \; &W^0, W^2, W^4 \in \mathbb R^{N_{\text{h}} \times N_{\text{i}}}; W^1, W^3 \in \mathbb R^{N_{\text{h}} \times N_{\text{h}}}; \\
				&W^5 \in \mathbb R^{N_{\text{o}} \times N_{\text{h}}}; b^0, b^1, b^2, b^3, b^4 \in \mathbb R^{N_\text{h}}; b^5 \in \mathbb R^{N_\text{o}} \},
			\end{aligned}
		\end{equation}
		where $N_{\mathrm{i}}$ is the number of units in the input layer, $N_{\mathrm{h}}$ the units per hidden layer, and $N_{\text{o}}$ the units in the output layer.
		In this demonstration, we use $N_{\mathrm{h}} = 10$, so $f_\NN \in \mathcal N(2 \to10 \to 1)$.
		Although NN efficiency or accuracy might improve with adjustments, this architecture is sufficient for all our demonstrations.
		Additional details on the choice of architecture and hyperparameters are provided in~\cref{sec:appendix2}.

		The initial condition is
		\begin{equation}
			q(x,0) = q_e(x,0) = \begin{cases}
				0 & 0 \leq x < 0.5 \\
				0.5 & x = 0.5 \\
				1 & 0.5 < x < 1,
			\end{cases}
		\end{equation}
		which is evolved to $t_f = 0.2$ for $N_t = 80$ time steps, when the exact solution is $q_e(x,t_f) = q_e(x-t_f,0)$.
		We define the loss as
		\begin{equation} \label{eqn:loss-scalar}
			\mathcal J = D(q, q_e) = \Delta x \sum_i \left[ q(x_i,t_f; \vec \theta\,) - q_e(x_i,t_f) \right]^2,
		\end{equation}
		which compares the solution mismatch only at the final simulation time $t_f$.
		Of course, intermediate time steps can be incorporated to the loss as needed.
		Weights and biases of $f_\NN$ are initialized with the Xavier initialization and trained by gradient descent using gradients computed with the full automatic differentiation (AD) provided by PyTorch~\cite{paszke2019pytorch}.
		Learning rates are adapted based on RMSprop~\cite{tieleman2012lecture} with the base learning rate of $10^{-3}$, the smoothing constant of $0.99$, and a small constant $\epsilon = 10^{-8}$ preventing division by zero.

		Results are illustrated in~\cref{fig:demo-advection}, labeled ``no constraint.''
		\Cref{fig:demo-advection}(b) shows that 1000 training iterations decrease $\mathcal J$ to about $10^{-3}$ of its initial value, suggesting that the model learns an operator consistent with the training data.
		However, the numerical solution in~\cref{fig:demo-advection}(a) oscillates rapidly near the sharp gradients, resembling typical artifacts from discretizing hyperbolic PDEs.
		There have been many efforts to develop numerical schemes that limit spurious oscillations for hyperbolic systems~\cite{leveque1992numerical}, yet they do not extend to NN training without additional development.

	\subsection{Constrained NN closure} \label{sec:constrainedNN}
		\subsubsection{Scalar advection in one dimension} \label{sec:constrainedNN-scalar}
		The method developed here for simple scalar advection generalizes to the more involved cases considered subsequently.
		A starting point is to recognize how \cref{eqn:discrete-scalar} also represents a finite-volume discretization of a scalar conservation law $\partial_t q + \partial_x f = 0$, where $u_i^n$ and $f_{i+\frac{1}{2}}^n$ would correspond to the piecewise constant representation of the solution and the approximate numerical flux:
		\begin{equation}
			q_i^n = \frac{1}{\Delta x} \int_{x_{i-\frac{1}{2}}}^{x_{i+\frac{1}{2}}} q(x,t^n) \, dx \quad \text{and} \quad f_{i+\frac{1}{2}}^n = \frac{1}{\Delta t} \int_{t^n}^{t^{n+1}} f(q(x_{i+\frac{1}{2}},t); \vec \theta\,) \, dt.
		\end{equation}
		Non-oscillatory finite-volume schemes preserve the solution smoothness by properly limiting the numerical flux, which requires further development for our NN fluxes.

		We start with Kurganov and Tadmor's (KT) central scheme~\cite{kurganov2000new}.
		This choice, which is made for simplicity, and other options are discussed in~\cref{sec:discussion}.
		Omitting time step indicator $n$ from the notation, the flux $f$ at $x_{i+\frac{1}{2}}$ is~\cite{kurganov2000new}
		\begin{equation} \label{eqn:nnflux}
			f_{i+\frac{1}{2}} = f^\mathrm{Rus}(q_{i+\frac{1}{2}}^+, q_{i+\frac{1}{2}}^-; \vec \theta\,),
		\end{equation}
		where $f^\mathrm{Rus}$ is the Rusanov flux for the NN $f_\NN$
		\begin{equation} \label{eqn:kt-scalar}
			f^\mathrm{Rus}(q_{i+\frac{1}{2}}^+,q_{i+\frac{1}{2}}^-; \vec \theta\,) = \frac{1}{2} \left[ f_\NN(q_{i+\frac{1}{2}}^+; \vec \theta\,) + f_\NN(q_{i+\frac{1}{2}}^-; \vec \theta\,) - a_{i+\frac{1}{2}}(\vec \theta\,) (q_{i+\frac{1}{2}}^+ - q_{i+\frac{1}{2}}^-) \right]
		\end{equation}
		with the local maximum wave speed
		\begin{equation} \label{eqn:wavespeed-scalar}
			a_{i+\frac{1}{2}}(\vec \theta\,) = \max \left[ \abs{f'_\NN(q_{i+\frac{1}{2}}^+; \vec \theta)}, \abs{f'_\NN(q_{i+\frac{1}{2}}^-; \vec \theta)} \right].
		\end{equation}
		For this illustration, $f_\NN \in \mathcal N(1 \to \cdot \to 1)$ is the needed single-input--single-output NN.
		Cell face values $q_{i+\frac{1}{2}}^\pm$ are the piecewise linearly reconstructed, slope-limited face states at $x_{i+\frac{1}{2}}$, from either the right $(x_{i+1})$ or the left $(x_i)$ volumes:
		\begin{equation} \label{eqn:reconstruction}
			q_{i+\frac{1}{2}}^+ = q_{i+1} - 0.5 \phi(r_{i+1}) (q_{i+2} - q_{i+1}) \quad \textrm{and} \quad q_{i+\frac{1}{2}}^- = q_i + 0.5 \phi(r_i) (q_{i+1} - q_i),
		\end{equation}
		where
		\begin{equation}\label{eqn:slope-ratio}
			r_i = \frac{q_i-q_{i-1}}{q_{i+1}-q_i}
		\end{equation}
		is the ratio of neighboring slopes, and $\phi(r) = \max[0,\min(1,r)]$ is the minmod limiter~\cite{sweby1984high}.
        The derivative of $f_\NN$ as in~\cref{eqn:wavespeed-scalar} can be computed exactly by AD evaluation of
		\begin{equation} \label{eqn:fprime-exact}
			f'_\NN(q_{i+\frac{1}{2}}^\pm; \vec \theta\,) = \pdv{q_{i+\frac{1}{2}}^\pm} \sum_j f_\NN(q_{j+\frac{1}{2}}^\pm; \vec \theta\,).
		\end{equation}

		\begin{figure}
            \centering
            \includegraphics[width=0.9\linewidth]{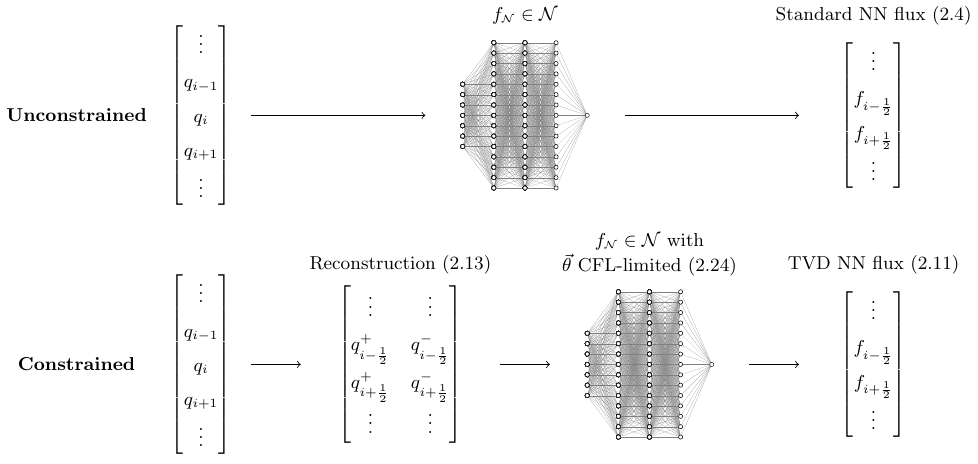}
			\caption{Schematic illustrating the difference between the unconstrained and the constrained NN closure.}
			\label{fig:nn-schematic}
		\end{figure}
		\Cref{fig:nn-schematic} illustrates how the constrained overall model~\cref{eqn:nnflux} can be regarded as constructed from the $\mathcal N$ model, augmented with additional operations on its input and output: the input is pre-processed by~\cref{eqn:reconstruction} and the output is post-processed by~\cref{eqn:kt-scalar}.
		The NN $f_\NN$ itself also needs to satisfy the Courant--Friedrichs--Lewy (CFL) condition while being trained, which will be discussed in~\cref{sec:optimization}, particularly~\cref{eqn:cfl-central}.

		What most directly provides the non-oscillatory property is the post-processing stage---the Rusanov flux~\cref{eqn:kt-scalar}---which, from the usual finite-volume perspective, adds numerical dissipation.
		The pre-processing---slope-limited reconstruction~\cref{eqn:reconstruction}---reduces the numerical dissipation, preserving the resolution.
		It is important to note that for the NN model, the Rusanov flux alone is insufficient to unconditionally constrain the model since the model parameters $\vec \theta$ must also be adjusted to satisfy the CFL constraint.
		Only together do these stages complete the strict enforcement.

		We designate the constrained version a ``TVD NN closure,'' where TVD indicates ``total variation diminishing,'' in the spirit of finite-volume TVD schemes~\cite{harten1997high}.
		In general, a numerical scheme or a numerical solution is called TVD if the total variation in space is non-increasing, with the total variation defined as
		\begin{equation} \label{eqn:tv-scalar}
			\mathrm{TV}[q] = \sum_i \abs{q_{i+1} - q_i}.
		\end{equation}
		We use ``TVD NN flux'' to refer to the flux $f$ in~\cref{eqn:nnflux}, constructed from $f_\NN$ in~\cref{eqn:kt-scalar}.
		The divergence of $f$ is embedded in the RHS of the PDE.
		As such, a TVD NN closure is a model composed of TVD NN fluxes.
		It will be demonstrated in~\cref{sec:demo} that the TVD NN closure does indeed preserve the TVD property.

	\subsubsection{Systems of equations in one dimension} \label{sec:constrainedNN-system}
		Extension to systems of equations is straightforward.
		The corresponding PDE system with $d$ state variables
		\begin{equation}
			\vec q_i^{\,n} = \begin{bmatrix}
				q_{1,i}^n \\ \vdots \\ q_{d,i}^n
			\end{bmatrix} = \begin{bmatrix}
				q_1(x_i, t^n) \\ \vdots \\ q_d(x_i, t^n)
		\end{bmatrix}
		\end{equation}
		evolves as
		\begin{equation} \label{eqn:discrete-system}
			\vec q_i^{\,n} = \vec q_i^{\,n-1} - \frac{\Delta t}{\Delta x} \left( \vec f_{i+\frac{1}{2}}^{\,n-1} - \vec f_{i-\frac{1}{2}}^{\,n-1} \right).
		\end{equation}
		Following standard developments~\cite{kurganov2000new}, the numerical flux $\vec f$ is defined again via the Rusanov flux $\vec f^{\,\mathrm{Rus}}$, again omitting the superscript $n$ for convenience:
		\begin{equation} \label{eqn:kt-system}
			\vec f_{i+\frac{1}{2}} = \vec f^{\,\mathrm{Rus}}(\vec q_{i+\frac{1}{2}}^{\,+}, \vec q_{i+\frac{1}{2}}^{\,-}; \vec \theta\,) = \frac{1}{2} \left[ \vec f_\NN(\vec q_{i+\frac{1}{2}}^{\,+}; \vec \theta\,) + \vec f_\NN(\vec q_{i+\frac{1}{2}}^{\,-}; \vec \theta\,) - a_{i+\frac{1}{2}}(\vec \theta\,) (\vec q_{i+\frac{1}{2}}^{\,+} - \vec q_{i+\frac{1}{2}}^{\,-}) \right],
		\end{equation}
		where
		\begin{equation} \label{eqn:wavespeed-system}
			a_{i+\frac{1}{2}}(\vec \theta\,) = \max \left[ \sigma \left( \pdv{\vec f_\NN}{\vec q} \, (\vec q^{\,+}_{i+\frac{1}{2}}) \, \right), \sigma \left( \pdv{\vec f_\NN}{\vec q} \, (\vec q^{\,-}_{i+\frac{1}{2}}) \, \right) \right].
		\end{equation}
		The $\vec f_\NN \in \mathcal (d \to \cdot \to d)$ is a NN model with $d$ inputs and $d$ outputs, and $\sigma(A)$ is the spectral radius of any matrix $A$.
		Computing the exact spectral radius of the Jacobian $\partial \vec f_\NN / \partial \vec q$ requires both constructing the matrix and computing its eigenvalues.
		As a simpler alternative, which especially simplifies the AD-based training, an upper bound of the spectral radius can be computed directly as
		\begin{equation}
			\sigma \left( \pdv{\vec f_\NN}{\vec q} \right) \leq \left\Vert \pdv{\vec f_\NN}{\vec q} \right\Vert_1 = \max_{1 \leq j \leq d} \sum_{i=1}^d \abs{\pdv{(f_{\NN})_i}{q_j}},
		\end{equation}
		with the 1-norm of the Jacobian now used for the nominal wave speed:
		\begin{equation} \label{eqn:wavespeed-system-approximate}
			a_{i+\frac{1}{2}}(\vec \theta\,) = \max \left[ \left\Vert \pdv{\vec f_\NN}{\vec q} \, (\vec q_{i+\frac{1}{2}}^{\,+} ) \, \right\Vert_1, \left\Vert \pdv{\vec f_\NN}{\vec q} \, (\vec q_{i+\frac{1}{2}}^{\,-}) \, \right\Vert_1 \right].
		\end{equation}
		The consequence of this estimate is only that the scheme is more dissipative.

		\subsubsection{Multiple dimensions}
			Extension to multiple dimensions follows by applying the one-dimensional schemes in each spatial direction.
			For example, in two dimensions with $\bf x = [x_1, x_2]^\top$, \cref{eqn:discrete-system} becomes
			\begin{equation}
				\vec q_{i,j}^{\,n} = \vec q_{i,j}^{\,n-1} - \frac{\Delta t}{\Delta x_1} (\vec f_{i+\frac{1}{2},j}^{\,n-1} - \vec f_{i-\frac{1}{2},j}^{\,n-1}) - \frac{\Delta t}{\Delta x_2} (\vec g_{i,j+\frac{1}{2}}^{\,n-1} - \vec g_{i,j-\frac{1}{2}}^{\,n-1}),
			\end{equation}
			where $\vec q_{i,j} = \vec q(x_{1,i},x_{2,j}) = \vec q(i \Delta x_1, j \Delta x_2)$, and $\vec f_{i+\frac{1}{2},j}$ and $\vec g_{i,j+\frac{1}{2}}$ are TVD NN fluxes evaluated on the cell faces at $(i+\frac{1}{2},j)$ and $(i,j+\frac{1}{2})$, respectively.
			The three-dimensional turbulent flame example in~\cref{sec:premixed} uses a straightforward extension of this.

	\subsection{Constrained training via projected gradient descent} \label{sec:optimization}

		As already discussed in~\cref{sec:constrainedNN-scalar}, a CFL condition is a key ingredient to avoid spurious oscillations:
		\begin{equation} \label{eqn:cfl-central}
			\max_{i,n} a_{i+\frac{1}{2}}^n(\vec \theta\,) \Delta t / \Delta x \leq \mathrm{CFL}_\mathrm{max},
		\end{equation}
		for all grid points $i$ and all time steps $n$ of the training.
		The maximum allowable value of the CFL number depends on the discretization and the time integration method.
		Central schemes---based on Nessyahu and Tadmor's scheme~\cite{nessyahu1990non}, including the KT scheme---in one dimension with forward Euler time stepping have $\mathrm{CFL}_\mathrm{max} =1/2$.

		A challenge is that $\max_{i,n} a_{i+\frac{1}{2}}^n(\vec \theta\,)$ varies as the NN parameters adjust each iteration as $\vec \theta^{\,k+1} = \vec \theta^{\,k} - \vec \alpha^k \odot (\mathrm{d} \mathcal J^k / \mathrm{d} \vec \theta\,)$.
		For an arbitrary $\Delta t$, it is probable that the maximum wave speed will exceed the limit in~\cref{eqn:cfl-central} during training.
		To prevent this and to achieve TVD, we define the feasible set of $\vec \theta$ as
		\begin{equation} \label{eqn:feasible-explicit}
			\mathcal C = \left\{ \vec \theta \ \Big\vert \ \max_{i,n} a_{i+\frac{1}{2}}^n(\vec \theta\,) \leq \bar a \equiv \mathrm{CFL}_\mathrm{max} \Delta x / \Delta t \right\},
		\end{equation}
		and the updated weights are mapped onto $\mathcal C$ every iteration,
		\begin{equation}
			\vec \theta^{\,k+1} = \pi_{\mathsmaller{\mathcal C}} \left( \vec \theta^{\,k} - \vec \alpha^k \odot \dv{\mathcal J^k}{\vec \theta} \right),
		\end{equation}
		ensuring that the model satisfies the constraint exactly.
		Since the spectral radius (or the 1-norm) of the Jacobian $\partial \vec f_\NN / \partial \vec q$ is linearly related to the output layer of the NN, rescaling its weights by a factor $\eta$ will also rescale the maximum wavespeed by $\eta$.
		The map $\pi_{\mathsmaller{\mathcal C}}$ is thus taken to be a simple rescaling of the weights of the output layer:
		\begin{equation} \label{eqn:map}
			\pi_{\mathsmaller{\mathcal C}}(\vec \theta) = \begin{cases}
				\eta W^l & l=5 \\
				W^l & l \neq 5
			\end{cases} \quad \textrm{where} \quad \eta = \begin{cases}
			1 & \max a(\vec q; W^5) \leq \bar a \\
			\eta': \max a(\vec q; \eta' W^5) = \bar a & \textrm{otherwise}.
			\end{cases}
		\end{equation}
		Here we omitted the subscript $i+\frac{1}{2}$ and the superscript $n$ for conciseness, yet $\max$ is still taken over $i$ and $n$.
		This map onto $\mathcal C$ is sufficient but not unique, which might warrant additional investigation.

		Finding the rescaling factor $\eta$ is straightforward when the solution $\vec q$ used to evaluate $a$ is independent of $\vec \theta$ so it does not change in the gradient descent iterations, such as for \textit{a priori} setup.
		In that case, $\max a(\vec q; \eta' W^5) = \eta' \max a(\vec q; W^5)$, so $\eta$ is exactly
		\begin{equation}
			\eta = \begin{cases}
				1 & \max a(\vec q; W^5) \leq \bar a \\
				\eta' = \bar a / \max a(\vec q; W^5) & \textrm{otherwise}.
			\end{cases}
		\end{equation}
		However, for \textit{a posteriori} training, $\vec q$ is nonlinear in $\vec \theta$, so $\eta'$ solves
		\begin{equation}
			\max a \big(\vec q(\eta' W^5); \eta' W^5 \big) = \bar a.
		\end{equation}
		Therefore, we find $\pi_{\mathsmaller{\mathcal C}}$ iteratively via~\cref{algo:iterative}.
		\begin{algorithm}
			\begin{algorithmic}
				\While{$\max a(\vec q; W^5) > \bar a$}
					\State $W^5 \gets [\bar a/\max a(\vec q; W^5)] W^5$ (rescale $W^5$)
					\State $\vec q \gets \vec q(W^5)$ (solve PDE again with the rescaled $W^5$)
				\EndWhile
			\end{algorithmic}
			\caption{Iterative rescaling procedure (only for \textit{a posteriori} setup)}
			\label{algo:iterative}
		\end{algorithm}

		A final remark is that the rescaling is not necessary if the time step size is small enough so the maximum wave speed never exceeds the upper limit $\bar a$.
		The rescaling only happens when the time step size is near the limit, and if it happens, the inner loop~\cref{algo:iterative} converges within 10 iterations for our demonstrations in~\cref{sec:advection} to~\cref{sec:antidiffusion}.
\section{Demonstrations} \label{sec:demo}

	\subsection{One-dimensional advection} \label{sec:advection}
        We first demonstrate the TVD NN model on the~\cref{sec:unconstrainedNN} case, with the NN trained to learn the RHS of the one-dimensional advection equation based on the optimization scenario~\cref{eqn:closure-scalar}.
        The flux $f$ is represented by the TVD NN~\cref{eqn:nnflux} with $f_\NN \in \mathcal{N}(1 \to 10 \to 1)$.
        Parameters are again Xavier initialized and optimized with RMSprop using the same hyperparameters as in~\cref{sec:unconstrainedNN}.

        \begin{figure}
        	\centering
        	\includegraphics[width=\linewidth]{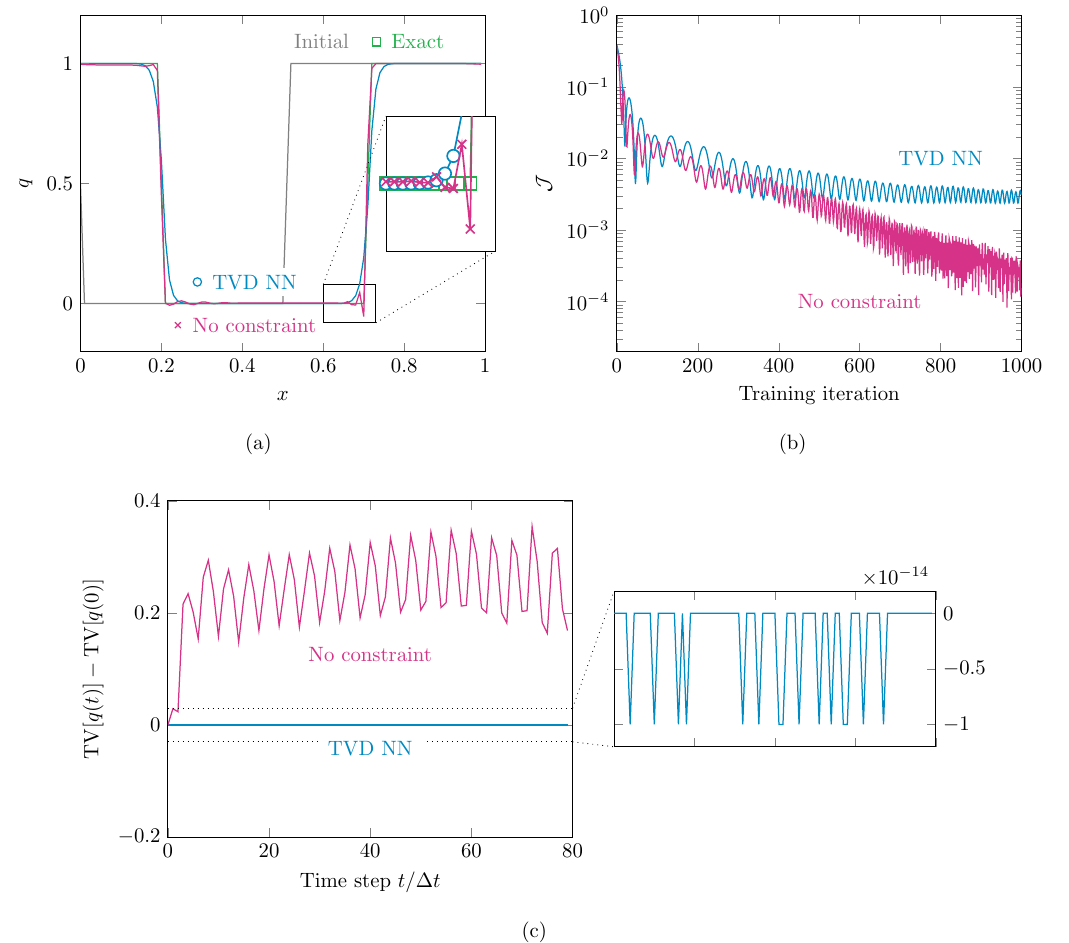}
        	\caption{Demonstration for the one-dimensional advection $\partial_t q + \partial_x q = 0$. (a) The unconstrained and the TVD NN predictions at $t=t_f$; (b) learning curves of both models; and (c) TV~\cref{eqn:tv-scalar} deviations from initial values.}
        	\label{fig:demo-advection}
        \end{figure}

    	\Cref{fig:demo-advection}(a) compares $q(x,t_f)$ numerical solutions for both cases with the exact solution.
    	Both successfully identify the underlying advection physics, which is confirmed by the learning curves in~\cref{fig:demo-advection}(b).
    	Of course, the strictly enforced constraint restricts the loss reducing capacity of the model, so the loss for the TVD NN is larger, but only the unconstrained NN introduces obvious spurious oscillations.
    	\Cref{fig:demo-advection}(c) shows that the TVD NN is, in fact, TVD with the TV value deviating from its initial value only by $\mathcal O(10^{-14})$, whereas the unconstrained model has qualitatively incorrect oscillations.

	\subsection{Burgers equation} \label{sec:burgers}
        Application to the inviscid Burgers equation $\partial_t q + \partial_x (q^2/2) = 0$ demonstrates extension to nonlinear dynamics.
        Again, the NN is required to learn the entire RHS, and $f_\NN \in \mathcal N(1 \to 10 \to 1)$ provides the TVD NN, with $f_\NN \in \mathcal N(2 \to 10 \to 1)$ for the corresponding unconstrained case.
        The discretization matches that used for the linear advection with 100 grid points on periodic $x \in [0,1]$ and 80 time steps until $t_f = 0.25$.
        The initial profiles for $q$ and the exact solution $q_e$ are discontinuous:
        \begin{equation}
        	q(x,0) = q_e(x,0) = \begin{cases}
        		1 & 0.375 \leq x < 0.625 \\ 0 & \text{otherwise},
        	\end{cases}
        \end{equation}
        with the exact solution
        \begin{equation}
        	q_e(x,t_f) = \begin{cases}
        		0 & 0 \leq x < 0.375 \\
        		4(x-0.375) & 0.375 \leq x < 0.625 \\
        		1 & 0.625 \leq x < 0.75 \\
        		0 & 0.75 \leq x < 1.
        	\end{cases}
        \end{equation}
        The loss is again defined as~\cref{eqn:loss-scalar}.
        RMSprop with the same hyperparameters as before is used for optimization.

		\begin{figure}
			\centering
			\includegraphics[width=\linewidth]{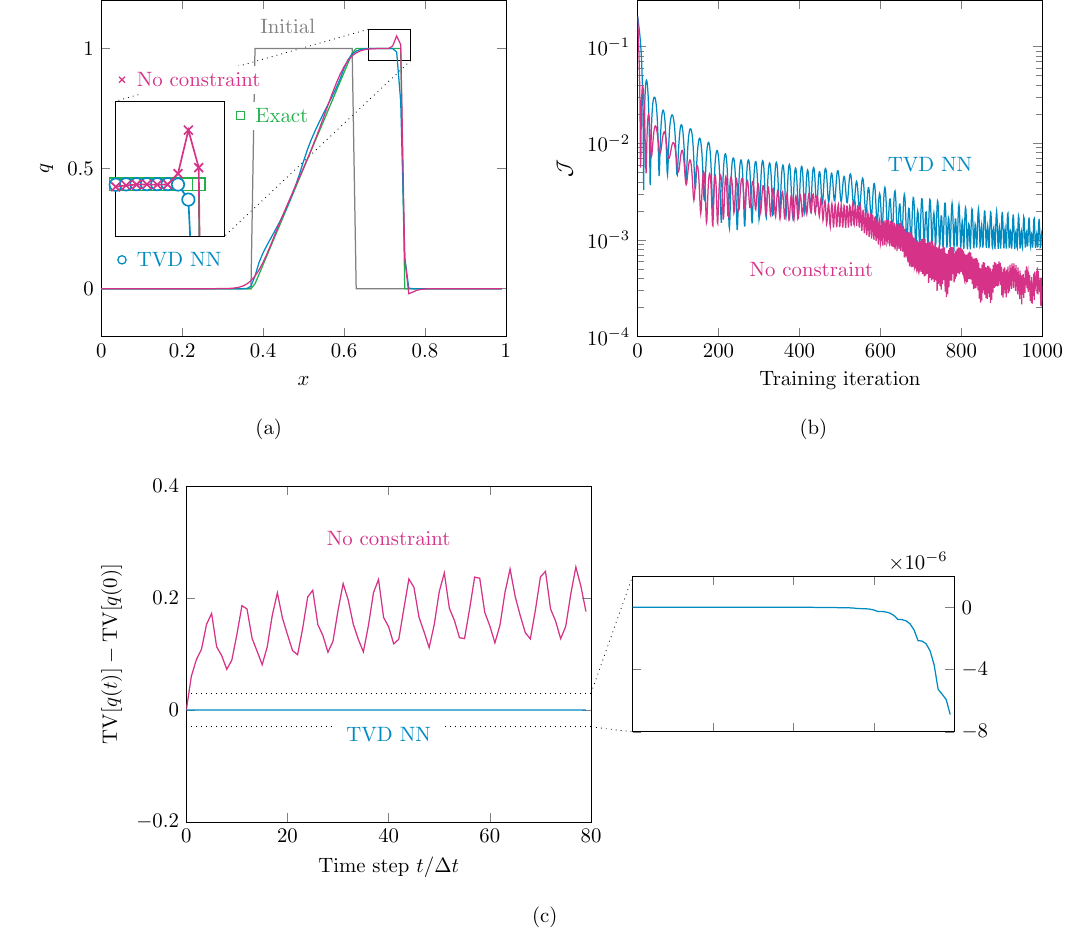}
			\caption{Demonstration for the one-dimensional Burgers equation $\partial_t q + \partial_x (q^2/2) = 0$. (a) Model predictions at $t=t_f$; (b) learning curves of both models; and (c) TV~\cref{eqn:tv-scalar} deviations from initial values.}
			\label{fig:demo-burgers}
		\end{figure}

		\Cref{fig:demo-burgers} shows that the TVD NN prediction is close to the exact solution without the spurious oscillations of the unconstrained NN.
		Again, the unconstrained model reaches a lower $\mathcal J$ but at the expense of introducing oscillations.
		In~\cref{fig:demo-burgers}(c), the TV value for the TVD NN decreases by a non-negligible amount after 60 time steps as the trained model overly dissipates the solution, though it only suggests that our model is TVD.

    \subsection{One-dimensional Euler equations} \label{sec:onedeuler}

	    In this case, a NN flux $\vec f$ from~\cref{eqn:discrete-system} is trained to learn the full RHS of the one-dimensional Euler equations based on its prediction of the conserved variables
		\begin{equation}
			\vec q = \begin{bmatrix}
				\rho \\ \rho u \\ E
			\end{bmatrix},
		\end{equation}
		where $\rho$ is the density, $u$ the velocity, and $E$ the total energy.
		The exact flux is
		\begin{equation} \label{eqn:euler}
			\vec f_e(\vec q) = \begin{bmatrix}
				\rho u \\ \rho u^2 + p \\ u(E+p)
			\end{bmatrix},
		\end{equation}
		with $\gamma = 1.4$ providing pressure $p = (\gamma-1) (E - \rho u^2/2)$.
		The model is trained to match the exact solution $\vec q_e$ with the loss defined as
		\begin{equation}
			\mathcal J = \Delta x \sum_i \left\Vert \vec q(x_i,t_f; \vec \theta\,) - \vec q_e(x_i,t_f) \right\Vert_2^2.
		\end{equation}
		We expect that $\vec f \to \vec f_e$ once $\mathcal J$ is minimized.

		The TVD property for a system of PDEs is more complex than the case of a single scalar equation.
		For example,
		\begin{equation} \label{eqn:tv-system}
			\mathrm{TV}[\vec q\,] = \sum_i \left \Vert \vec q_{i+1} - \vec q_i \right \Vert_1
		\end{equation}
        is insufficient to check the TVD property of a scheme because the exact solution $\vec q_e$ itself may have increasing total variation.
        For a system, one way is to measure the total variation of the amplitude of characteristic waves.
        This can be done analytically for certain PDEs, including the Euler equations; however, in general, the decomposition can be cumbersome~\cite[Section 15]{levy1999central}.
        As a result, we do not monitor the exact TVD property of the scheme for systems, though the benefits will be obvious.

		The spatial domain $x \in [0,1]$ is discretized in the same manner as the scalar case with 501 mesh points, including boundary points, yielding $\Delta x = 2 \times 10^{-3}$.
		Time is discretized as $\Delta t = 10^{-4}$ over $t \in [0.1, 0.15]$.
		The model prediction $\vec q$ is initialized at $t=0.1$ with $\vec q_e(x,t=0.1)$ and we compare the solutions at $t_f = 0.15$.
		The exact solution $\vec q_e$ follows the Sod configuration~\cite{sod1978survey} with the initial condition
		\begin{equation}
			\vec q_e(x,0) = \begin{cases}
				\vec q_{\mathsmaller L} & x < 0.5 \\ \vec q_{\mathsmaller R} & x \geq 0.5
			\end{cases}, \quad \vec q_{\mathsmaller L} = \begin{bmatrix} 1.0 \\ 1.0 \\ 0.0 \end{bmatrix}, \quad \vec q_{\mathsmaller R} = \begin{bmatrix} 0.125 \\ 0.1 \\ 0.0 \end{bmatrix}.
		\end{equation}
		Early $t \in [0,0.1]$ is not used for training to focus on the richer behavior that follows the step initial condition.
		In particular, testing for $t \in [0,0.1]$ yields excellent suppression of oscillations but poor learning of physics underlying the moving discontinuities.

		The solution features do not reach the boundaries of the domain, so homogeneous Neumann boundary conditions are sufficient at $x = 0$ and $1$.
        The constrained NN model is $\vec f_\NN \in \mathcal{N}(3 \to 50 \to 3)$ while the unconstrained model is also based on the formulation~\cref{eqn:discrete-system} but with $\vec f_{i+\frac{1}{2}} = \vec f_\NN(\vec q_i, \vec q_{i+1}; \vec \theta\,) \in \mathcal N(6 \to 50 \to 3)$.
        Again, RMSprop is used for the training with the same hyperparameter values as before.
        All the weights except for the output layer $W^5$ are Xavier initialized; $W^5$ is initialized as zero to prevent an unnecessary overshoot of $\mathcal J$ in early gradient descent iterations.

		\begin{figure}
			\centering
			\includegraphics[width=0.9\linewidth]{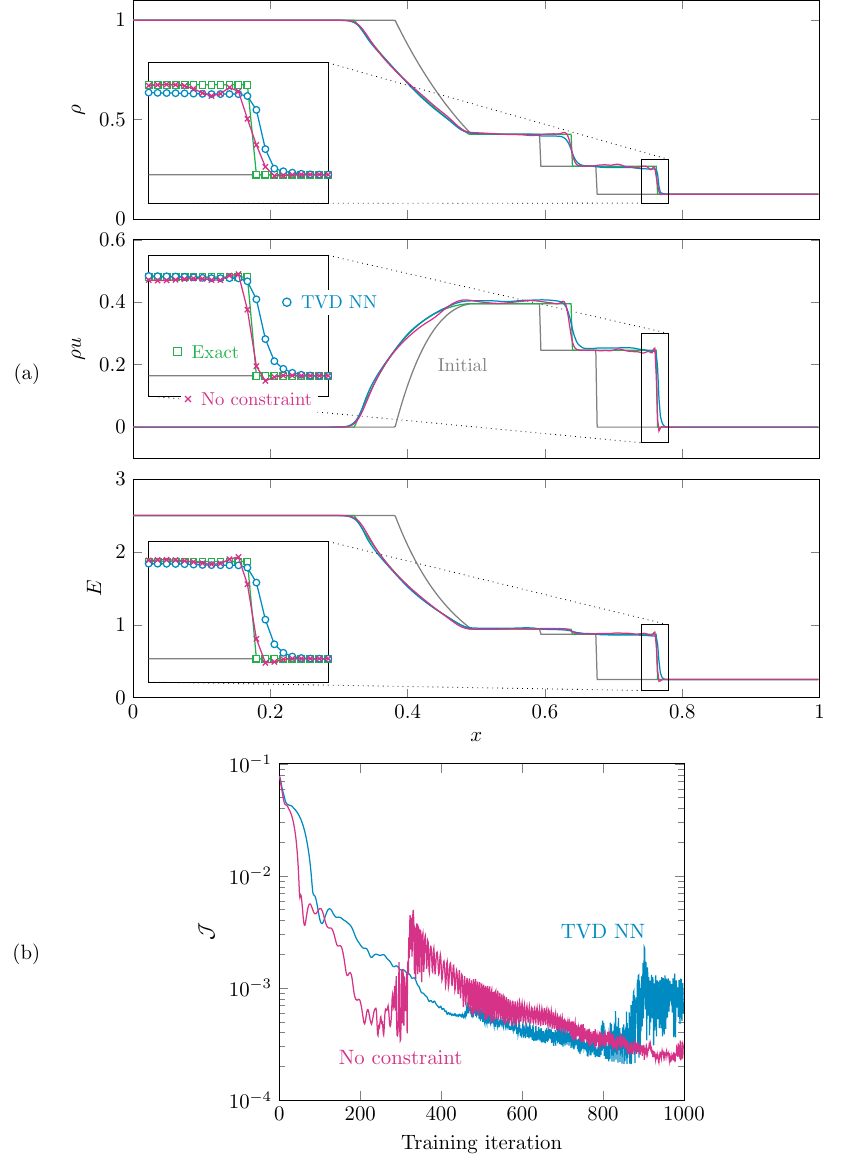}
			\caption{Demonstration for the one-dimensional Euler equations. (a) Numerical solutions of $\vec q(x,t_f)$ predicted by both models, initial conditions, and the analytical solutions; (b) learning curves of both models.}
			\label{fig:demo-euler}
		\end{figure}

		\Cref{fig:demo-euler} shows that both NN models successfully reproduce the gross features, as might be expected for a similar finite-volume scheme, reducing the loss significantly, but the TVD NN also suppresses oscillations.
		Loss oscillating at $\mathcal J \lesssim 10^{-3}$ for both cases is expected as the model nears a local minimum and RMSprop starts searching more broadly around it.

\subsection{One-dimensional advection with anti-diffusion} \label{sec:antidiffusion}
		Thus far, only hyperbolic systems have been considered with clear TVD properties.
		A final example, before the application to turbulent combustion, is not hyperbolic.
		This requires a modification of the formulation.
		Since $f_\NN$ in~\cref{eqn:kt-scalar} and $\vec f_\NN$ in~\cref{eqn:kt-system} are local, depending only on pointwise grid values, both are only capable of learning hyperbolic behaviors.
		It cannot represent a diffusion-like behavior, which is more fundamentally nonlocal and can, therefore, be expected to require multi-point input like a finite difference gradient.
		We generalize the model design to address this.

		The most convenient example would be advection-diffusion; however, significant diffusion so readily suppresses oscillation that a more illustrative scenario involves an anti-diffusive flux that triggers instabilities.
		Anti-diffusion is also relevant to turbulent flames.
		For example, consider the filtered transport equation of a flame progress variable $c$ for LES:
		\begin{equation} \label{eqn:ctransport-filtered}
			\begin{split}
				\pdv{\ol \rho \wt c}{t} &= -\nabla \cdot (\ol \rho \wt{\bf u c}) - \nabla \cdot (\ol \rho \wt{\bm \varphi}_c) + \ol{\dot \omega}_c \\
				&= -\nabla \cdot (\ol \rho \wt{\bf u} \wt c) - \nabla \cdot (\ol \rho \wt{\bm \varphi}_c) + \ol{\dot \omega}_c - \nabla \cdot (\ol \rho \bm{\varphi}^r_c),
			\end{split}
		\end{equation}
		where $\bf u$ is the flow velocity, $\dot \omega_c$ the production rate of $c$ by chemical reaction, and $\bm \varphi_c$ the diffusive flux.
		Given a variable $q$, $\ol q = \mathcal F q$ and $\wt q = \mathcal F \rho q / \mathcal F \rho$ denote its filtered and Favre-filtered values, respectively, by a filter $\mathcal F$.
		For this illustration, only the sub-filter-scale convection $\bm \varphi^r_c = \wt{\bf u c} - \wt{\bf u} \wt c$ is considered in~\cref{eqn:ctransport-filtered}; filtered terms involving $\wt{\bm \varphi}_c$ and $\ol{\dot \omega}_c$ are left unclosed.
		A common gradient-diffusion model is $\bm \varphi^r_c = -\mathcal D_T \nabla \wt c$	with the eddy diffusivity $\mathcal D_T$.
		However, this assumes that $\bm \varphi^r_c$ is aligned with $\nabla \wt c$, but near a flame, $\bm \varphi^r_c$ can be counter to the gradient~\cite{veynante1997gradient,macart2018effects}, which warrants an anti-diffusive flux.
		A NN closure model for $\bm \varphi^r_c$ will be most effective if it is capable of learning anti-diffusion.
		Of course, unconstrained anti-diffusion introduces spurious oscillations, leading eventually to unbounded $\wt c$.
		In combustion, violations of $\wt c \in [0,1]$ as seen in~\cref{fig:example}, where $Y = 1-\wt c$, are qualitatively incorrect, and similar out-of-bound behaviors can have challenging consequences for more complex physiochemical models.

\begin{figure}
			\centering
			\includegraphics[width=0.8\linewidth]{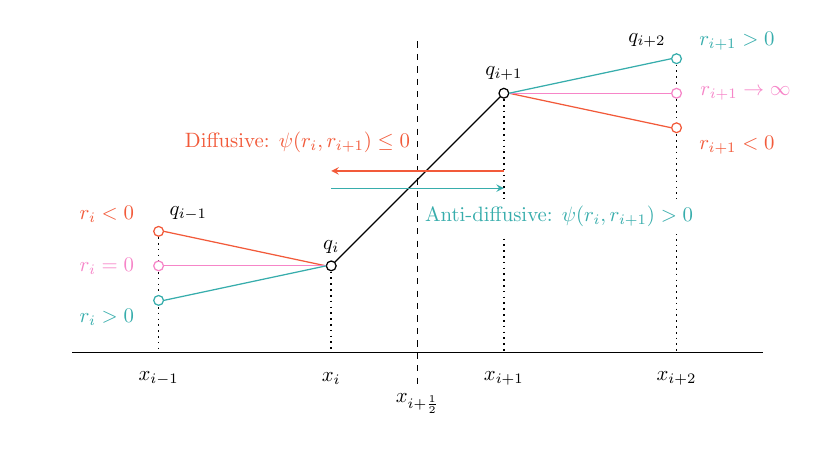}
			\caption{Action of the limiter for the anti-diffusive flux. Arrows indicate the direction of the $x_{i+\frac{1}{2}}$ flux.}
			\label{fig:antidiffusion}
		\end{figure}

		Based on~\cref{eqn:nnflux}, the numerical flux for a one-dimensional scalar is generalized as
		\begin{equation} \label{eqn:constrained+diffusion}
			f_{i+\frac{1}{2}} = \frac{1}{2} \left[ f_\NN(q_{i+\frac{1}{2}}^+; \vec \theta_f) + f_\NN(q_{i+\frac{1}{2}}^-; \vec \theta_f) - a_{i+\frac{1}{2}}(\vec \theta_f) (q_{i+\frac{1}{2}}^+ - q_{i+\frac{1}{2}}^-) \right] - \hat \nu_{i+\frac{1}{2}} \frac{q_{i+1}-q_i}{\Delta x},
		\end{equation}
		where $\hat \nu_{i+\frac{1}{2}}$ is the NN-predicted local diffusivity.
		To limit the anti-diffusive contribution of $\hat \nu$, it is decomposed into two parts:
		\begin{equation}
			\hat \nu_{i+\frac{1}{2}} = \abs{\nu_\NN^+ (q_i,q_{i+1}; \vec \theta_{\nu^+})} + \psi(r_i, r_{i+1}) \nu_\NN^- (q_i,q_{i+1}; \vec \theta_{\nu^-}),
		\end{equation}
		where $\abs{\nu_\NN^+}$ and $\nu_\NN^-$ represent the diffusive and potentially anti-diffusive components, respectively.
		Both $\nu_\NN^+$ and $\nu_\NN^-$ are $\in \mathcal N(2 \to \cdot \to 1)$.
		While $\nu_\NN^+$ is cast positive as $\abs{\nu_\NN^+}$ so its effect is strictly diffusive, $\nu_\NN^-$ is not, but $\psi$ limits its anti-diffusive effect based on the shape of the solution:
		\begin{equation}
			\psi(r_i,r_{i+1}) = \begin{cases}
				\min \big[ \min(r_i,1/r_i), \min(r_{i+1},1/r_{i+1})\, \big] & \min(r_i, r_{i+1}) > 0 \\
				0 & \textrm{otherwise},
			\end{cases} \label{eqn:limiter-antidiffusion}
		\end{equation}
		where $r_i$ is the same ratio of neighboring slopes defined in~\cref{eqn:slope-ratio}.
		For this demonstration, a small number $\epsilon_r = 10^{-12}$ is added to the denominator to prevent division by zero as
		\begin{equation} \label{eqn:slope-ratio+eps}
			r_i = \frac{q_i - q_{i-1}}{q_{i+1} - q_i + \epsilon_r}.
		\end{equation}
		\Cref{fig:antidiffusion} sketches the mechanism of the flux limiter: anti-diffusion is permitted only when the sequence $(q_{i-1},q_i,q_{i+1},q_{i+2})$ is monotone---when both $r_i$ and $r_{i+1}$ are positive.
		If either $q_i$ or $q_{i+1}$ is a local extremum, $\psi(r_i,r_{i+1})$ is zero, suppressing anti-diffusion.
		Anti-diffusivity $\psi \nu_\NN^-$ is maximized when $r_i=r_{i+1}=1$, which corresponds to a locally linear profile.

		The formulation is demonstrated for a solution with flame-like evolution from a smooth profile to a sharp discontinuity while being advected at a constant speed.
		A time-reversed heat equation solution is used to construct the target solution for training.
		We define $w$ the solution of $\partial_t w = \nu \partial_{xx} w$ with $\nu = 0.01$ with the initial condition
		\begin{equation}
			w(x,0) = \begin{cases}
				0 & 0 \leq x < 0.5 \\ 1 & 0.5 \leq x < 1
			\end{cases}
		\end{equation}
		on a periodic domain $x \in [0,1]$.
		The target solution is $q_e(x,t) = w(x-t, t_f-t)$, which is the solution of the advection-anti-diffusion equation $\partial_t q + \partial_x q = - \nu \partial_{xx} q$ that evolves from smooth $w(x,t_f)$ at $t=0$ to sharp $w(x-t_f,0)$ at $t=t_f$.
		The NNs $f_\NN$, $\nu_\NN^+$, and $\nu_\NN^-$ are trained simultaneously to minimize
		\begin{equation}
			\mathcal J = \Delta x \sum_i \left[ q(x_i,t_f; \vec \theta\,) - q_e(x_i,t_f) \right]^2,
		\end{equation}
		where $q(x,0) = q_e(x,0)$ and $\vec \theta = [\vec \theta_f; \vec \theta_{\nu^+}; \vec \theta_{\nu^-}]$.
		Here, the~\cref{sec:unconstrainedNN} discretization for both space and time is used with $t_f = 0.2$.

		The NN models are $f_\NN \in \mathcal N(1 \to 10 \to 1)$ and $\nu_\NN^\pm \in \mathcal N(2 \to 10 \to 1)$, along with an unconstrained NN $f_\NN \in \mathcal N(2 \to 10 \to 1)$ for comparison.
		A TVD NN closure model based on~\cref{eqn:kt-scalar} without the $\hat \nu$ modification is also included for comparison.
		All models are trained using RMSprop with the same hyperparameters, and model parameters are Xavier initialized.
		The raw output of $\nu_\NN^\pm$ is multiplied by the physical viscosity $\nu$ to provide a good initial guess to the training predictions, which helps prevent overshoots in early training iterations.

\begin{figure}
			\centering
			\includegraphics[width=\linewidth]{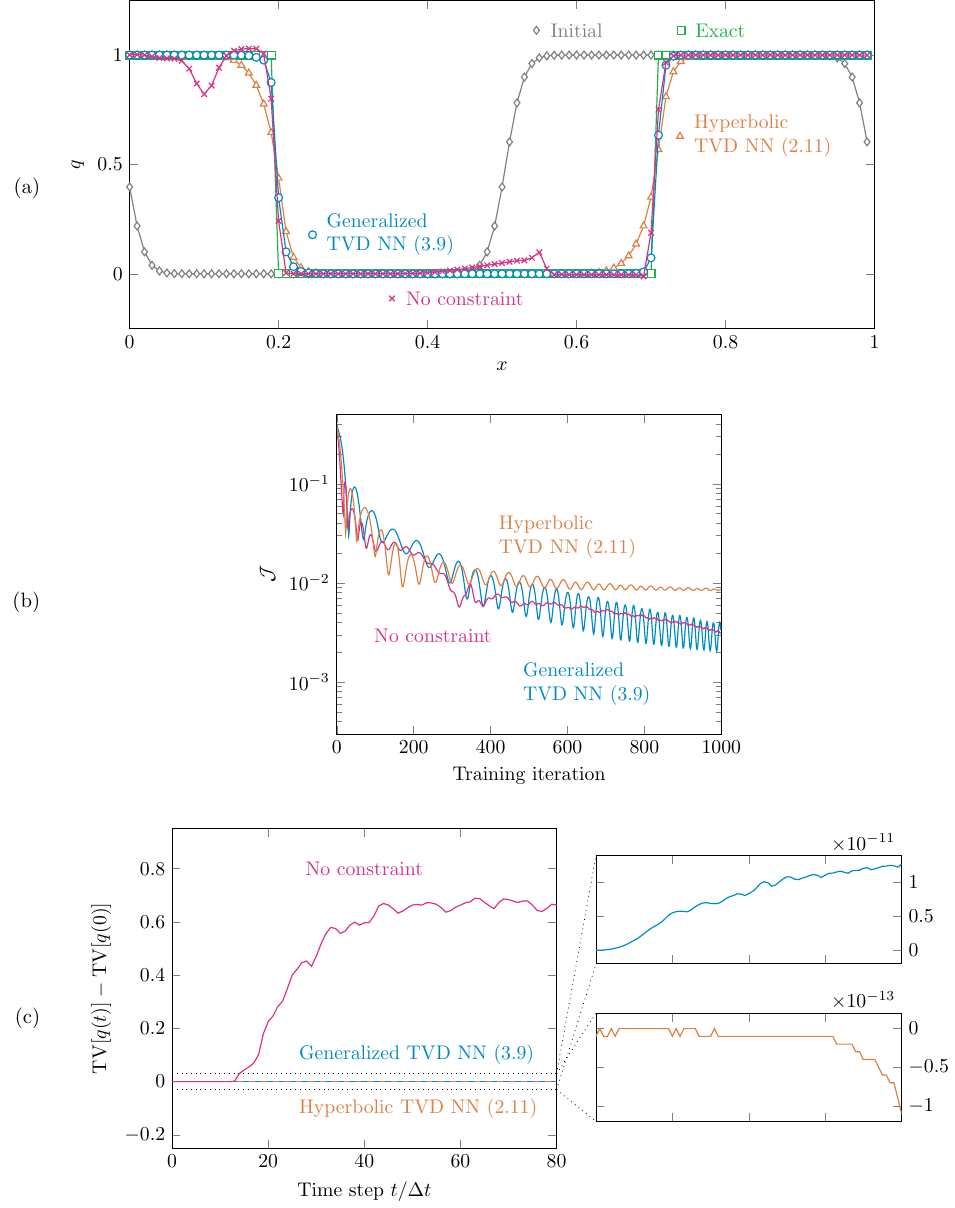}
			\caption{One-dimensional advection with anti-diffusion: (a) numerical solutions at $t=t_f$, the exact solution, and the initial condition; (b) learning curves; and (c) TV~\cref{eqn:tv-scalar} deviations from initial values.}
			\label{fig:demo-antidiffusion}
		\end{figure}

		Numerical solutions predicted by all models are plotted in~\cref{fig:demo-antidiffusion}(a), and the learning curves and TV histories are plotted in~\cref{fig:demo-antidiffusion}(b) and~\cref{fig:demo-antidiffusion}(c).
		We observe that the constrained model with $\hat \nu$ successfully predicts the displaced and sharpened profile without the obvious oscillations of the unconstrained model.
		The TV value for the generalized model~\cref{eqn:constrained+diffusion} increases but only slightly, which is due to the error introduced in~\cref{eqn:slope-ratio+eps}.
		Thus, we regard it bounded $q \in [0,1]$ within tolerance as desirable for a flame progress variable.
		Oscillations in the unconstrained model~\cref{eqn:nn-unconstrained} prediction are qualitatively different from typical spurious oscillations in hyperbolic problems.
		They even appear away from sharp gradients due to the anti-diffusion instability: in~\cref{fig:demo-antidiffusion}(c), the instability becomes obvious after around 15 time steps.
		The hyperbolic-only model~\cref{eqn:kt-scalar} fails to learn effective anti-diffusion, showing that the nonlocal generalization of~\cref{eqn:constrained+diffusion} is beneficial.

    \subsection{Premixed flame in three-dimensional turbulence} \label{sec:premixed}

		Finally, the TVD NN is applied to SGS modeling for the flame in a turbulent premixture of~\cref{fig:example}.
		The simulation is adapted from Towery \textit{et al.}~\cite{towery2016spectral}.
		The nondimensional reacting compressible Navier--Stokes equations
		\begin{equation} \label{eqn:gov-flame-nondim}
			\partial_t \vec q + \nabla \cdot \vec{\bf f}(\vec q\,) = \vec s(\vec q\,)
		\end{equation}
		for
		\begin{equation}
			\vec q = \begin{bmatrix}
				\rho \\ \rho \bf u \\ E \\ \rho Y
			\end{bmatrix},
		\end{equation}
		govern the flow, in a three-dimensional space $\bf x = [x_1, x_2, x_3]^\top$ and flow velocity $\bf u = [u_1, u_2, u_3]^\top$.
		The chemical reaction is taken to be single-species, single-step, and irreversible for the reactant mass fraction $Y$: a single reactant is perfectly converted to a product past the flame, so $Y=1$ for reactant, and $Y=0$ for the product.
		$E$ is the sum of the internal and kinetic energy, not including the chemical energy.
		The transport flux and the non-conservative source term are
		\begin{equation}
			\vec{\bf f}(\vec q\,) = \begin{bmatrix}
				\rho \bf u \\ \rho \bf u \bf u + p \bf I \ \\ \bf u (E+p) \\ \rho \bf u Y
			\end{bmatrix} - \begin{bmatrix}
				0 \\ \bm \tau \\ \bm \tau \cdot \bf u - \bm \varphi_T \\ -\bm \varphi_Y
			\end{bmatrix} \quad \textrm{and} \quad \vec s(\vec q\,) = \begin{bmatrix}
				0 \\ \bf 0 \\ Q \dot \omega \\ - \dot \omega
			\end{bmatrix}.
		\end{equation}
		The viscous stress and diffusive fluxes are
		\begin{equation}\label{eqn:flame-flux-nondim}
			\begin{gathered}
				\bm \tau = \frac{\mu}{\mathrm{Re}} \left[ \nabla \bf u + (\nabla \bf u)^\top - \frac{2}{3} (\nabla \cdot \bf u) \bf I \right], \quad
				\bm \varphi_T = -\frac{\mu}{\mathrm{Re\,Pr}} \nabla T, \quad \textrm{and} \quad
				\bm \varphi_Y = -\frac{\mu}{\mathrm{Re\,Sc}} \nabla Y,
			\end{gathered}
		\end{equation}
		where Re is the Reynolds number, Pr the Prandtl number, and Sc the Schmidt number.
		The viscosity depends on the temperature $T$ as $\mu = (T/T_0)^{0.7}$ with $T_0$ the unburnt gas temperature, and the ideal gas equation of state is $p = (\gamma-1) \rho T / \gamma$, where the pressure is $p = (\gamma-1) (E - \rho \bf u \cdot \bf u / 2 )$.
		The chemical reaction rate is $ \dot \omega = A \rho^2 Y \exp[ (\beta/\alpha) (1-T_f/T) ]$, in which $A$ is the Arrhenius constant, $\alpha$ the heat release ratio, $\beta$ the Zel'dovich number, $T_f = T_0/(1-\alpha)$ the adiabatic flame temperature, and $Q = \alpha T_f$ the heat release parameter.
		See~\cref{sec:nondim} for more details, including the nondimensionalization and the parameter values.

		The governing equations are filtered and solved for the filtered state variables
		\begin{equation}
			\filter \vec q (\bf x,t) = \int F_\filtersize(\bf x-\bf x') \vec q(\bf x',t) \, d\bf x',
		\end{equation}
		where $F_\filtersize$ is a spatial filter kernel with its filter width $\mathlarger{\filtersize}$ defining the LES, and $\filter$ is the corresponding filter operator.
		The governing equations for $\filter \vec q$ are
		\begin{equation}
			\partial_t (\filter \vec q\,) = -\nabla \cdot (\filter \vec{\bf f}(\vec q\,)) + \filter \vec s(\vec q\,) = -\nabla \cdot [\vec{\bf f}(\filter \vec q\,) + \vec{\bf f}^r_e] + [\vec s(\filter \vec q\,) + \vec s^{\,r}_e],
		\end{equation}
		where the exact unclosed residuals are
		\begin{equation}
			\vec{\bf f}^r_e(\vec q\,) = \filter \vec{\bf f}(\vec q\,) - \vec{\bf f}(\filter \vec q\,) \quad \text{and} \quad \vec s^{\,r}_e(\vec q\,) = \filter \vec s(\vec q\,) - \vec s(\filter \vec q\,).
		\end{equation}
		Both $\vec{\bf f}^r_e$ and $\vec s^{\,r}_e$ need to be closed based on $\filter \vec q$, so the closure modeling task is to find
		\begin{equation} \label{eqn:closure-problem}
			\vec{\bf f}^r(\filter \vec q\,) \approx \vec{\bf f}^r_e(\vec q\,) \quad \text{and} \quad \vec s^{\,r}(\filter \vec q\,) \approx \vec s^{\,r}_e(\vec q\,)
		\end{equation}
		approximations.
		Hereafter, we only consider the filtered variables and omit $\filter$ for conciseness.
		A variable $q$ denotes either $\filter q$ if $q \in \{ \rho, p \}$ or $\filter \rho q / \filter \rho$ if $q \in \{ \bf u, T, Y \}$.

		The SGS models $\vec{\bf f}^r$ and $\vec s^{\,r}$ are compositions of elementary NNs $\in \mathcal N$ and TVD NNs, trained toward~\cref{eqn:closure-problem}.
		We represent them as
		\begin{equation} \label{eqn:nn-flame-const}
			\vec{\bf f}^r(\vec q; \vec \theta_f) = \begin{bmatrix}
				\bf 0 \\
				\bf f_{\bf u} \\
				\bf f_{\bf u} \cdot \bf u + \rho \bf f_T \\
				\rho \bf f_Y
			\end{bmatrix} \quad \text{and} \quad \vec s^{\,r}(\vec q; \vec \theta_s) = \begin{bmatrix}
				0 \\ \bf 0 \\ Q s_\NN \\ -s_\NN
		\end{bmatrix} \dot \omega (\vec q\,),
		\end{equation}
		where
		\begin{equation}
			\bf f_{\bf u} = \begin{bmatrix}
				f_{\NN,11}(\vec q; \vec \theta_{f_{11}}) & f_{\NN,12}(\vec q; \vec \theta_{f_{12}}) & f_{\NN,13}(\vec q; \vec \theta_{f_{13}}) \\
				f_{\NN,21} & f_{\NN,22}(\vec q; \vec \theta_{f_{22}}) & f_{\NN,23}(\vec q; \vec \theta_{f_{23}}) \\
				f_{\NN,31} & f_{\NN,32} & f_{\NN,33}(\vec q; \vec \theta_{f_{33}})
			\end{bmatrix}
		\end{equation}
		closes the momentum equations.
		It is composed of 6 independent NNs $\in \mathcal N$, where the symmetry is enforced by letting $f_{\NN,ji} = f_{\NN,ij}$ for $(i,j) \in \{ (1,2),(1,3),(2,3) \}$.
		This is to keep the model $\bf f_{\bf u}$ consistent with the exact residual $\filter(\rho \bf u \bf u) - (\filter \rho) (\filter \bf u) (\filter \bf u)$, which is a symmetric tensor.

		Fluxes
		\begin{equation}
			\bf f_T = \begin{bmatrix}
				f_{T1} \\ f_{T2} \\ f_{T3}
			\end{bmatrix} \quad \text{and} \quad \bf f_Y = \begin{bmatrix}
				f_{Y1} \\ f_{Y2} \\ f_{Y3}
			\end{bmatrix}
		\end{equation}
		close the energy and species equations, where $f_{Ti}$ and $f_{Yi}$ are taken to be coupled.
		That is, each pair $(f_{Ti}, f_{Yi})$ is the output from a single TVD NN:
		\begin{equation}
			\vec f_{c1}(\vec q; \vec \theta_{c1}) = \begin{bmatrix}
				f_{T1} \\ f_{Y1}
			\end{bmatrix}, \quad
			\vec f_{c2}(\vec q; \vec \theta_{c2}) = \begin{bmatrix}
				f_{T2} \\ f_{Y2}
			\end{bmatrix}, \quad \text{and} \quad
			\vec f_{c3}(\vec q; \vec \theta_{c3}) = \begin{bmatrix}
				f_{T3} \\ f_{Y3}
			\end{bmatrix}.
		\end{equation}
		The TVD NN in~\cref{eqn:constrained+diffusion} is extended to a system of equations and is used for $\vec f_{ci}$ as
		\begin{equation}
			\begin{split}
				\vec f_{ci}(\vec q; \vec \theta_{ci}) = \frac{1}{2} \left[ \vec f_{\NN,ci} (\vec q_c^{\,i+}; \vec \theta_{f_{ci}}) + \vec f_{\NN,ci}(\vec q_c^{\,i-}; \vec \theta_{f_{ci}})  - a(\vec \theta_{f_{ci}}) (\vec q_c^{\,i+} - \vec q_c^{\,i-}) \right]& \\
				- \left[ \abs{\vec \nu_{\NN,ci}^{\,+}(\vec q; \vec \theta_{\nu^+_{ci}})} + \vec \psi_i(\vec q_c) \odot \vec \nu_{\NN,ci}^{\,-}(\vec q; \vec \theta_{\nu^-_{ci}}) \right] \odot \frac{\Delta_i \vec q_c}{\Delta x_i}&,
			\end{split} \label{eqn:nnflux-reactive-scalars}
		\end{equation}
		forming $\vec \theta_{ci} = [\vec \theta_{f_{ci}}; \vec \theta_{\nu_{ci}^+}; \vec \theta_{\nu_{ci}^-}]$, for $i \in \{1,2,3\}$.
		The symbol $\odot$ indicates component-wise multiplication.
		In~\cref{eqn:nnflux-reactive-scalars}, $\vec q_c = [\hat T, Y]^\top$ is a vector of reactive scalars with the normalized temperature $\hat T = T/T_0$, $\vec q_c^{\,i\pm}$ the slope-limited reconstruction of $\vec q_c$, $\vec \psi_i$ the anti-diffusive flux limiter~\cref{eqn:limiter-antidiffusion} for each component of $\vec q_c$, and $\Delta_i \vec q_c / \Delta x_i$ the finite-differenced gradient of $\vec q_c$, all in the $x_i$-direction.
		The $T$-component of the raw output of $\vec f_{ci}$, $f_{Ti}$, is multiplied by $T_0$ for normalization.
		More details of the NN formulation can be found in~\cref{sec:nnform}.

		The NN SGS stress $\bf f_{\bf u}$ for the momentum is not constrained because, without bounds corresponding to $Y \in [0,1]$, the lack of smoothness of the velocity field is less consequential.
		That is, there is nothing qualitatively incorrect for any small error in $\bf u$.
		To retain consistency with the momentum equations, $\bf f_{\bf u} \cdot \bf u = \{ f_{\NN,ij} u_j\}_{i=1}^3$ is added to the total energy equation.
		Finally, the NN correction for the non-conservative source term $s_\NN$ is added to the equations for the reactive scalars, which is multiplied by $\dot \omega(\vec q\,)$ so it becomes active only near the flame where $\dot \omega(\vec q\,) \not \approx 0$.
		The NN $s_\NN$ also does not require constraint, although its effect on the smoothness of the solution is not yet fully explored.
		All NN models in~\cref{eqn:nn-flame-const}---$f_{\NN,ij}$ for momentum, $\vec f_{\NN,ci}$ and $\vec \nu_{\NN,ci}^{\,\pm}$ for scalars, and $s_\NN$---share the baseline architecture:
		\begin{equation}
			f_{\NN,ij}, s_\NN \in \mathcal N(39 \to 50 \to 1), \quad \vec f_{\NN,ci} \in \mathcal N(2 \to 50 \to 2), \quad \vec \nu_{\NN,ci}^{\,\pm} \in \mathcal N(39 \to 50 \to 2)
		\end{equation}
		for $i \in \{1, 2, 3\}$ and $j \geq i$.

		To generate a training dataset, DNS is run on a rectangular domain as in~\cref{fig:flame-schematic}, discretized as a uniform structured staggered mesh with $256 \times 256 \times 4096$ points.
		A standard fourth-order Runge--Kutta (RK4) method is used---given a PDE $\partial_t \vec q = \vec R(\vec q)$,
		\begin{equation} \label{eqn:rk4}
			\begin{split}
				\vec q^{\,n,1} &= \vec q^{\,n,0} + \frac{\Delta t}{2} \vec R(\vec q^{\,n,0}) \\
				\vec q^{\,n,2} &= \vec q^{\,n,0} + \frac{\Delta t}{2} \vec R(\vec q^{\,n,1}) \\
				\vec q^{\,n,3} &= \vec q^{\,n,0} + \Delta t \vec R(\vec q^{\,n,2}) \\
				\vec q^{\,n+1,0} = \vec q^{\,n,4} &= \vec q^{\,n,0} + \Delta t \sum_{s=1}^4 w_s \vec R(\vec q^{\,n,s-1}),
			\end{split}
		\end{equation}
		for $\vec q^{\,n} = \vec q^{\,n,0}$ at time step $n$ with the weights $w_s = [\frac{1}{6}, \frac{1}{3}, \frac{1}{3}, \frac{1}{6}]^\top$.
		Then, the DNS solutions are box-filtered
		\begin{equation}
			F_\filtersize(\bf x) = \begin{cases}
				1/\mathlarger{\filtersize}^3 & \Vert \bf x \Vert_\infty \leq \mathlarger{\filtersize}/2 \\
				0 & \textrm{otherwise}
			\end{cases},
		\end{equation}
		with the filter size of $\mathlarger{\filtersize}/\Delta_\textsc{dns} = 16$, which is then downsampled to the LES grid of $\Delta_\textsc{les} = \mathlarger{\filtersize}$ with the grid size of $16 \times 16 \times 256$.

		For every gradient-descent iteration, a minibatch size of $M=20$ is randomly sampled from the total $\mathcal M = 1000$ DNS snapshots.
		LES predictions are initialized with the downsampled filtered DNS field at a time stamp $t_0^{(m)}$ of the $m$-th snapshot for $m \in \{1, 2, \ldots, M\}$.
		The NN model leads to the LES solution $\vec q$ by numerically integrating
		\begin{equation} \label{eqn:les+ml}
			\partial_t \vec q + \nabla \cdot [\vec{\bf f}(\vec q\,) + \vec{\bf f}^r(\vec q;\vec \theta_f)] = [\vec s(\vec q\,) + \vec s^{\,r}(\vec q;\vec \theta_s)]
		\end{equation}
		on $t \in [t_0^{(m)}, t_0^{(m)}+t_f]$ with RK4~\cref{eqn:rk4} for some simulation time $t_f$.
		Then, following previous efforts with embedded SGS NN models~\cite{sirignano2020dpm,macart2021embedded}, the parameters
		\begin{equation}
			\vec \theta = \left[\{ \vec \theta_{f_{ij}} \}_{i=1,j \geq i}^3; \{\vec \theta_{f_{ci}}\}_{i=1}^3; \{\vec \theta_{\nu_{ci}^\pm}\}_{i=1}^3; \vec \theta_s \right]
		\end{equation}
		are trained \textit{a posteriori}, to minimize the averaged loss
		\begin{equation} \label{eqn:loss-flame}
			\mathcal J = \frac{1}{M} \sum_{m=1}^M \mathcal J^{(m)} = \frac{1}{M} \sum_{m=1}^M \int_\Omega \left\Vert \vec q_p(\bf x, t_0^{(m)}+t_f; \vec \theta\,) - \vec q_{p,e}(\bf x, t_0^{(m)}+t_f) \right\Vert_W^2 \, d \bf x,
		\end{equation}
		which measures the mismatch between the NN-predicted solution of primitive variables $\vec q_p \equiv [\rho, \bf u, \hat T, Y]^\top$ and the trusted data $\vec q_{p,e}$---downsampled filtered DNS data---within a region of interest $\Omega$ at time $t_0^{(m)}+t_f$ for all $m \in \{ 1, 2, \ldots, M \}$.
		We use a weighted norm
		\begin{equation}
			\Vert \vec v \Vert_W^2 = \vec v^\top W \vec v \quad \text{with} \quad W = \mathrm{diag}(0,1,1,1,3,3),
		\end{equation}
		which normalizes the mismatch of each variable, where the coefficients are selected based on predictions.
		The mismatch of $\rho$ is not included since the Favre-filtered continuity equation does not require closure, though including it might warrant investigation.
		The simulation time $t_f$ is selected to be short enough that the sensitivity does not explode~\cite{chung2022optimization}.
		We choose $t_f=0.1$ which corresponds to 200 LES time steps with $\Delta t_\textsc{les} = 5 \times 10^{-4}$.

		For comparison, an unconstrained NN model is tested where
		\begin{equation} \label{eqn:nn-flame-unconst}
			\vec{\bf f}^r(\vec q; \vec \theta_f) = \begin{bmatrix}
				\bf 0 \\ \bf f_{\bf u} \\ \bf f_{\bf u} \cdot \bf u + \bf f_T \\ \bf f_Y
			\end{bmatrix} \quad \text{and} \quad \vec s^{\,r}(\vec q; \vec \theta_s) = \begin{bmatrix}
				0 \\ \bf 0 \\ Q s_\NN \\ -s_\NN
			\end{bmatrix} \dot \omega(\vec q\,),
		\end{equation}
		which matches~\cref{eqn:nn-flame-const} but all entries of $\bf f_T = \{ f_{\NN,Ti} \}_{i=1}^3$ and $\bf f_Y = \{ f_{\NN,Yi} \}_{i=1}^3$ are unconstrained NN models $\in \mathcal N(39 \to 50 \to 1)$.
		Again, raw outputs of $f_{\NN,Ti}$ are multiplied by $T_0$ for normalization.

		A penalized model using~\cref{eqn:nn-flame-unconst} is also tested for comparison, where the bounded $Y \in [0,1]$ property is sought by penalizing the loss as
		\begin{equation} \label{eqn:loss+penalty}
			\mathcal J = \frac{1}{M} \sum_{m=1}^M \left[ \mathcal J^{(m)} + \int_\Omega P(Y(\bf x,t_0^{(m)}+t_f;\vec \theta\,)) \, d\bf x \right]
		\end{equation}
		with
		\begin{equation}
			P(Y) = \begin{cases}
				Y^2 & Y < 0 \\
				(Y-1)^2 & Y > 1 \\
				0 & \text{otherwise},
			\end{cases}
		\end{equation}
		which penalizes out-of-bound values.

		All NN parameters are Xavier initialized, while the weights of the output layer $W^5$ are initialized as zero to prevent overshoots.
		Adam~\cite{kingma2014adam} is used as the optimizer, with the base learning rate of $10^{-3}$ and $L_2$ regularization with the regularization constant of $\lambda = 10^{-2}$.
		Additional details of the numerical simulations, the NN formulation, and the optimization are in~\cref{sec:appendix}.

\begin{figure}
			\centering
			\includegraphics[width=\linewidth]{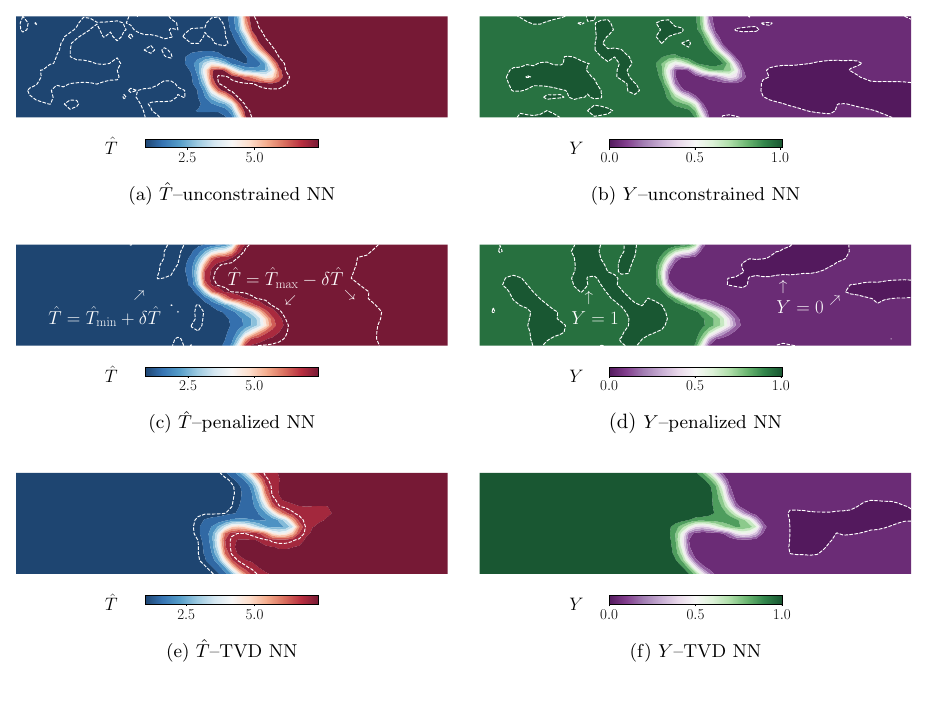}
			\caption{Turbulent flame LES: (a) normalized temperature $\hat T = T/T_0$ and (b) reactant mass fraction $Y$ for the unconstrained model~\cref{eqn:nn-flame-unconst}; (c) $\hat T$ and (d) $Y$ for the model~\cref{eqn:nn-flame-unconst} with penalization~\cref{eqn:loss+penalty}; and (e) $\hat T$ and (f) $Y$ for the TVD NN model~\cref{eqn:nn-flame-const}. Dashed contours in (a), (c), and (e) indicate where $\hat T = \hat T_\mathrm{max} - \delta \hat T$ or $\hat T = \hat T_\mathrm{min} + \delta \hat T$ with $\delta \hat T = 10^{-2} (\hat T_\mathrm{max} - \hat T_\mathrm{min})$; those in (b), (d), and (f) indicate where $Y=0$ or $Y=1$: enclosed regions are out of bound $Y \not \in [0,1]$. The bound $Y \leq 1$ is exactly satisfied in (f).}
			\label{fig:flame}
		\end{figure}

\begin{figure}
			\centering
			\includegraphics[width=0.5\linewidth]{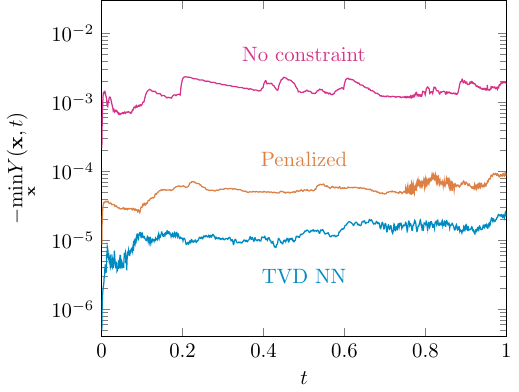}
			\caption{Violations of the $Y \geq 0$ constraint: magnitude of minimum $Y$ values for all cases.}
			\label{fig:minY-flame}
		\end{figure}

		The TVD NN method far outperforms the others.
		\Cref{fig:flame} shows $\hat T$ and $Y$ predicted by the unconstrained, penalized, and TVD NN models on a two-dimensional slice of the computational domain.
		The prediction is out of sample in the sense that it is made at $t=1$, 2000 time steps from the initial flow field (about one large-eddy turnover time), which is well beyond the 200 time steps ($t_f = 0.1$) used for training.
		In this case, the unconstrained and penalized NNs both lead to unphysical fluctuations in both $\hat T$ and $Y$ fields, which are most obvious away from the flame where both should be uniform.
		Although there is no strict constraint on $\hat T$ that it can fluctuate at large wavelengths as pressure waves are resolved, small-scale oscillations near $\hat T_\mathrm{min}$ for the unconstrained and penalized models are unphysical.
		The TVD NN successfully suppresses the oscillations in both scalar fields, except one $Y<0$ region, for which potential reasons will be discussed in~\cref{sec:premixed-discussion}.
		Still, the violation is milder than that for the other models, as in~\cref{fig:minY-flame}.


\section{Additional Discussion} \label{sec:discussion}

	This section further discusses the approach and its results, particularly for the turbulent combustion problem, including additional factors to consider and potential opportunities.

	\subsection{Comments on the turbulent flame results from~\cref{sec:premixed}} \label{sec:premixed-discussion}
	It was seen in~\cref{fig:flame} that the non-oscillatory (boundedness) property is significantly improved, but we emphasize it may not be exactly preserved.
	There are several potential reasons for this, which warrant discussion.

	First, the NN terms are coupled with the existing physical source terms.
	Physical solutions can oscillate, as can discretizations of them, irrespective of the added NN closure.
	If the numerical scheme used for the represented terms allows oscillations, then the solution is not guaranteed to be non-oscillatory since it is not fully represented with the constrained NN.
	There is only an expectation that the added NN closure itself will not contribute to them.
	Specifically, while conservative flux divergence terms in~\cref{eqn:gov-flame-nondim} can be constrained by upwinding the flux as in~\cref{sec:flame-numerics}, there is no simple way to exactly constrain non-conservative source terms to preclude oscillations.
	Of course, there exist standard approaches for this issue~\cite{bermudez1994upwind}, but their generalization to our reacting Navier--Stokes equations is cumbersome.
	Moreover, the effect of the source term closure $s_\NN$ may exacerbate the inexactness of the physical source term, which has not yet been considered and is left for future work.

	Second, the performance of the multi-dimensional extension of the scheme has not been considered in detail.
	Although the common practice for the extension is to apply a one-dimensional scheme in each dimension, the full consequences are unexplored, and there are possibly additional performance opportunities.

	Lastly, the effect of using the standard RK4~\cref{eqn:rk4} is unknown since the TVD property of our model has only been verified for the forward Euler scheme.
	A safer choice is a TVD time-stepper that preserves TVD property, such as the Strong Stability Preserving Runge--Kutta (SSP RK) methods~\cite{gottlieb1998total}.
	However, it should be noted that the TVD time steppers preserve such property only when used in conjunction with an exact TVD numerical flux.
	Since only the NN closure terms are made TVD, not the whole discretization to avoid dissipation that degrades turbulence, there is no significant advantage anticipated for a TVD time scheme.
	We used~\cref{eqn:rk4} as it is accurate and convenient, unlike SSP RK methods that require auxiliary operators for more than third-order accurate schemes~\cite{gottlieb1998total}.
	Moreover, the adjoint of~\cref{eqn:rk4} needed to compute sensitivities $\mathrm{d} \mathcal J / \mathrm{d} \vec \theta$ for~\cref{eqn:loss-flame} has been derived previously~\cite{vishnampet2015practical}.
	Despite such benefits and the success seen in~\cref{fig:flame}, its potential inexactness still needs further investigation.

	\subsection{Added computational cost of TVD NN} \label{sec:performance}
	Here, we evaluate the cost of the limiting procedure itself, which is most important with our focus on preventing violations assuming a suitable NN is otherwise available as discussed in~\cref{sec:intro}.
	\Cref{table:cost} shows the cost of TVD and non-TVD NN closures for all of our demonstrations, both for evaluating and training the models.
	The constrained model is overall less than 10 times more expensive than the unconstrained model for all scenarios.
	This might seem excessive.
	However, we emphasize that the non-TVD models are qualitatively incorrect.
	In cases where that is catastrophic to a calculation, the TVD NN cost is worth the cost.
	Therefore, the main conclusion is that the TVD NN approach is not prohibitive, allowing the use of a NN closure that would otherwise fail.

	\begin{table}
		\centering
		\scalebox{0.9}{
			\begin{tabular}{c | c c | c c}
				\hline \hline
				\multirow{2}{*}{Scenario} & \multicolumn{2}{c|}{Training (sec/gradient descent step)} & \multicolumn{2}{c}{Evaluation (sec/time step)} \\
				\cline{2-5}
				& Unconstrained & Constrained & Unconstrained ($\times 10^{-3}$) & Constrained ($\times 10^{-3}$) \\
				\hline
				\Cref{sec:advection} & 0.029 & 0.11 & 0.23 & 1.0 \\ 
				\Cref{sec:burgers} & 0.028 & 0.11 & 0.23 & 1.2 \\
				\Cref{sec:onedeuler} & 1.2 & 3.4 & 0.64 & 2.1 \\
				\Cref{sec:antidiffusion} & 0.042 & 0.28 & 0.38 & 2.8 \\
				\Cref{sec:premixed} & 500 & 980 & 640 & 1200 \\
				\hline \hline
			\end{tabular}
		}
		\caption{Wall time elapsed (s) to train or evaluate NN closure models in each scenario. Training means the entire gradient descent iterations with both forward and backward sweeps, and evaluation means a single PDE solve. All one-dimensional cases (\cref{sec:advection} to \cref{sec:antidiffusion}) are run on a single CPU core (Intel i5), and the flame case (\cref{sec:premixed}) is run on a single GPU (NVIDIA V100).}
		\label{table:cost}
	\end{table}	

	\subsection{Alternate base schemes} \label{sec:alternatives}

	We adapted the KT scheme to NN for its efficiency compared to most other upwind schemes, which require characteristic decomposition for application to systems of equations.
	This generally requires iterative methods, which make the differentiation (by hand or with AD) needed more complex.
	Central schemes avoid this, and the KT scheme is one of the most concise among them.
	That said, there is no impediment for other established schemes including upwind schemes to be adapted to the new NN closure.

	As a central scheme, the model can be further modified, especially in the reconstruction step.
	The slope-limited reconstruction~\cref{eqn:reconstruction} is not necessary for the non-oscillatory property but for minimizing the numerical dissipation and improving accuracy.
	For that, one could employ different slope limiters, as the KT scheme~\cite{kurganov2000new} is compatible with other options including the monotonized central limiter $\phi(r) = \max[0, \min(2r, 0.5(r+1), 2)]$.
	Additional reconstruction methods can also be adapted, such as Weighted ENO (WENO) reconstruction~\cite{jiang1996efficient}.
	Although WENO reconstruction is typically used for upwind schemes, it can also be applied to central schemes~\cite{levy1999central}, so it should be compatible with the specific formulation introduced.
	Note that for higher-order reconstruction for systems, additional benefits should come from computing the exact spectral radius of the NN flux Jacobian~\cref{eqn:wavespeed-system} instead of estimating it from the~\cref{eqn:wavespeed-system-approximate} 1-norm, although the needed eigenvalue decomposition would make differentiation more challenging.

	Finally, one can replace the differentiation~\cref{eqn:fprime-exact} with a less expensive approximation, which can potentially reduce the cost that has been evaluated in~\cref{table:cost}.
	As $f'_\NN$ is a function of $\vec \theta$ and affects the loss, the NN training requires the gradient of $f'_\NN$ with respect to $\vec \theta$.
	The gradient $\mathrm d \mathcal J / \mathrm d \vec \theta$ includes the $\partial_{f'_\NN} \mathcal J \partial_{\vec \theta} f'_\NN$ component from the chain rule, where $\partial_{\vec \theta} f'_\NN$ is then a second-order derivative of $f_\NN$, $\partial_{\vec \theta} f'_\NN = \partial_{\vec \theta} \, \partial_{q^\pm} f_\NN(q^\pm)$, which increases the cost of NN training.
	Finite difference is an alternative,
	\begin{equation}
		f'_\NN(q_{i+\frac{1}{2}}^\pm) \approx \frac{f_\NN(q_{i+1}) - f_\NN(q_i)}{q_{i+1} - q_i},
	\end{equation}
	but this is challenging when $q_{i+1} \approx q_i$, so we chose~\cref{eqn:fprime-exact}, which is more expensive but exact.

	\subsection{General applicability}	
	We have formulated the model, particularly~\cref{eqn:constrained+diffusion}, so it can learn both hyperbolic and parabolic PDEs.
	Extending the discussion from the first paragraph of~\cref{sec:antidiffusion}, this was important for a NN closure to be capable of learning the needed physics.
	We anticipate that the modified model covers a wide class of conservation laws with suitable closures for unknown or unrepresented physics.
	For learning higher-order dynamics, either providing a wider stencil to the NNs or adding more complexities to the NN itself should suffice.
	Including fully nonlocal effects, such as forms of radiation, would necessitate significant redesign.
	We can anticipate that the proposed model may be challenged if asked to learn PDEs with non-conservative source terms since the formulation is in a conservative form.
	In this scenario, an additional NN source term should probably be introduced, as in~\cref{eqn:nn-flame-const}, though the investigation of its effect on the non-oscillatory property is left for future work.

	A final remark is that the TVD NN is not guaranteed to stay exactly constrained when applied to out-of-sample data.
    Recall that $\vec \theta$ is constrained within the feasible set in \cref{eqn:feasible-explicit}, and the maximum wave speed is evaluated only over the training samples.
    That is,
	\begin{equation}
		\mathcal C = \left\{ \vec \theta \ \Big\vert \max_{q \in \mathcal Q_\mathrm{train}} a(\vec \theta) \Big\vert_q \leq \mathrm{CFL}_\mathrm{max} \Delta x / \Delta t \right\},
	\end{equation}
	where $\mathcal Q_\mathrm{train}$ is the training dataset of $q$.
	Hence, if the trained model is evaluated on $q \not \in \mathcal Q_\mathrm{train}$, there is no strict guarantee that the predicted solution is non-oscillatory.
	The hope is that $q$ is similar enough to some $q' \in \mathcal Q_\mathrm{train}$ that the model preserves this property well enough.
	In other words, inclusive $\mathcal Q_\mathrm{train}$ will provide robustness.
	This suggests that training a model \textit{a posteriori} is likely more robust than training it \textit{a priori}, since $\mathcal Q_\mathrm{train} = \{ q(\bf x,t) \,\vert\, t \in [t_0^{(m)},t_0^{(m)}+t_f] \}_{m=1}^M$ for the \textit{a posteriori} training in~\cref{eqn:loss-flame}, for example, is more abundant than $\mathcal Q_\mathrm{train} = \{ q(\bf x,t_0^{(m)})\}_{m=1}^M$ for the corresponding \textit{a priori} training, though the proposed scheme is not limited to either of the approaches.

\section*{Acknowledgments}
This material is based in part upon work supported by the Department of Energy, National Nuclear Security Administration, under Award Number DE-NA0003963.
\section*{CRediT Authorship Contribution Statement}
\begin{description}
	\item \textbf{Seung Won Suh:} Conceptualization, methodology, software, validation, formal analysis, investigation, data curation, writing -- original draft, visualization.
	\item \textbf{Jonathan F. MacArt:} Conceptualization, methodology, software, writing -- review \& editing, supervision.
	\item \textbf{Luke N. Olson and Jonathan B. Freund:} Conceptualization, methodology, writing -- review \& editing, supervision, project administration, funding acquisition.
\end{description}
\section*{Supplementary Materials}
A repository with codes used for numerical experiments from~\cref{sec:advection} to~\cref{sec:antidiffusion} is published for reproducibility: \url{https://github.com/swsuh28/demo_tvd_nn}.
\appendix
\section{Additional Details for Turbulent Flame Simulations} \label{sec:appendix}

	\subsection{Nondimensionalization} \label{sec:nondim}
		We derive~\cref{eqn:gov-flame-nondim} from its dimensional form:
		\begin{equation}
			\pdv{t^\star} \begin{bmatrix}
				\rho^\star \\ \rho^\star \bf u^\star \\ E^\star \\ \rho^\star Y
			\end{bmatrix} + \nabla^\star \cdot \begin{bmatrix}
				\rho^\star \bf u^\star \\
				\rho^\star \bf u^\star \bf u^\star + p^\star \bf I - \bm \tau^\star \\
				\bf u^\star (E^\star + p^\star) - \bm \tau^\star \cdot \bf u^\star + \bm \varphi^\star_T \\
				\rho^\star \bf u^\star Y + \bm \varphi^\star_Y
			\end{bmatrix} = \begin{bmatrix}
				0 \\ \bf 0 \\ Q^\star \dot \omega^\star \\ -\dot \omega^\star
			\end{bmatrix},
		\end{equation}
		where $\star$ denotes a dimensional quantity.
		By definition, the reactant mass fraction $Y$ is already dimensionless.
		Constitutive relations for diffusive fluxes are
		\begin{equation} \label{eqn:flame-flux-dim}
			\bm \tau^\star = \mu^\star \left[ \nabla^\star \bf u^\star + (\nabla^\star \bf u^\star)^\top - \frac{2}{3} (\nabla^\star \cdot \bf u^\star) \bf I \right], \quad \bm \varphi^\star_T = -k^\star \nabla^\star T^\star, \quad \bm \varphi^\star_Y = -\rho^\star \mathcal D^\star \nabla^\star Y,
		\end{equation}
		where diffusivities are $T^\star$ dependent as
		\begin{equation}
			\frac{\mu^\star}{\mu^\star_0} = \frac{k^\star/c_p^\star}{k^\star_0/c_{p,0}^\star} = \frac{\rho^\star \mathcal D^\star}{\rho^\star_0 \mathcal D^\star_0} = \left( \frac{T^\star}{T^\star_0} \right)^{0.7}.
		\end{equation}
		The subscript $0$ denotes the unburnt state, and we assume a constant specific heat capacity at constant pressure, so $c_p^\star = c_{p,0}^\star$.
		The ideal gas equation of state is
		\begin{equation}
			p^\star = \rho^\star \mathcal R_\mathrm{sp}^\star T^\star,
		\end{equation}
		with the specific gas constant $\mathcal R_\mathrm{sp}^\star$, and the Arrhenius law
		\begin{equation}
			\dot \omega^\star = A^\star \rho^{\star 2} Y \exp(-T_a^\star/T^\star),
		\end{equation}
		with the Arrhenius constant $A^\star$ and the activation temperature $T_a^\star$.

		We use the length scale $\ell^\star_\scf$ and the velocity scale $u^\star_\scf$ from the forcing scheme discussed in~\cref{sec:forcing}, the unburnt gas density $\rho^\star_0$, and the specific heat capacity $c_p^\star$ to nondimensionalize.
		The resulting nondimensional variables are
		\begin{equation}
			\bf x = \frac{\bf x^\star}{\ell^\star_\scf}, \quad \bf u = \frac{\bf u^\star}{u^\star_\scf}, \quad t = \frac{t^\star u^\star_\scf}{\ell^\star_\scf}, \quad \rho = \frac{\rho^\star}{\rho^\star_0}, \quad p = \frac{p^\star}{\rho^\star_0 u^{\star 2}_\scf}, \quad \text{and} \quad T = \frac{c_p^\star T^\star}{u_\scf^{\star 2}},
		\end{equation}
		and we define $T_0 = c_p^\star T_0^\star / u_\scf^{\star 2}$ the nondimensional reference temperature.
		Diffusive fluxes~\cref{eqn:flame-flux-dim} become~\cref{eqn:flame-flux-nondim}, as we define
		\begin{equation}
			\mathrm{Re} = \frac{\rho_0^\star u_\scf^\star \ell_\scf^\star}{\mu_0^\star}, \quad \mathrm{Pr} = \frac{\mu_0^\star c_p^\star}{k_0^\star}, \quad \text{and} \quad \mathrm{Sc} = \frac{\mu_0^\star}{\rho_0^\star \mathcal D_0^\star},
		\end{equation}
		with $\mu = \mu^\star / \mu_0^\star = (T/T_0)^{0.7}$.
		The equation of state becomes $p = (\gamma-1) \rho T / \gamma$, and the Arrhenius law becomes
		\begin{equation}
			\dot \omega = \underbrace{A^\star \rho_0^{\star 2} \exp(-T_a^\star/T_f^\star)}_{\equiv A} \rho^2 Y \exp(\frac{T_a^\star}{T_f^\star} - \frac{T_a^\star}{T^\star}) = A \rho^2 Y \exp[ \frac{\beta}{\alpha} \left( 1 - \frac{T_f}{T} \right) ],
		\end{equation}
		with $\alpha = (T_f^\star - T_0^\star) / T_f^\star$ and $\beta = \alpha T_a^\star / T_f^\star$, where $T_f^\star$ is the dimensional adabatic flame temperature.

		Parameter values in~\cref{table:param} are adapted from Towery \textit{et al.}~\cite{towery2016spectral}.
		The reference temperature $T_0$ is elevated to lower the Mach number $\mathrm{Ma} = 1/\sqrt{(\gamma-1)T_0}$ to prevent unwanted transition to detonation following Poludnenko  \& Oran~\cite{poludnenko2011spontaneous}.
		\begin{table}
			\centering
			\begin{tabular}{c c c}
				\hline
				Reynolds number & Re & 2500 \\
				Prandtl number & Pr & 0.1 \\
				Schmidt number & Sc & 0.1 \\
				Unburnt gas temperature & $T_0$ & 2353 \\
				Arrhenius constant & $A$ & 2500 \\
				Heat release ratio & $\alpha$ & 0.863 \\
				Zel'dovich number & $\beta$ & 5.49 \\
				Heat capacity ratio & $\gamma$ & 1.17 \\
				\hline
			\end{tabular}
			\caption{Dimensionless parameters.}
			\label{table:param}
		\end{table}

	\subsection{Numerical procedures} \label{sec:flame-numerics}
		\begin{figure}
			\centering
			\includegraphics[width=\linewidth]{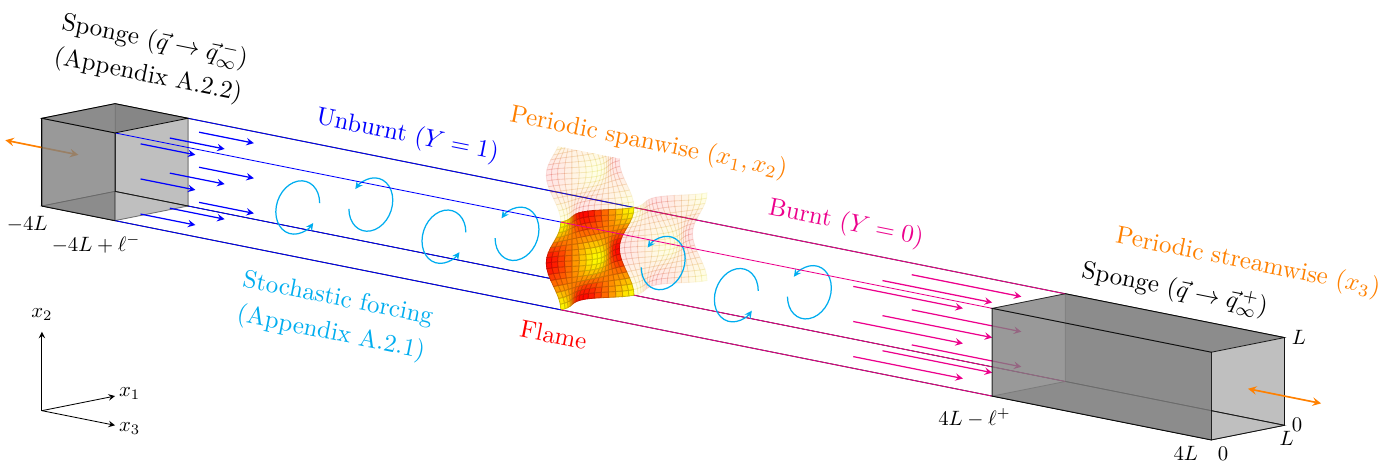}
			\caption{Schematic of the turbulent flame simulation.}
			\label{fig:flame-schematic}
		\end{figure}
		\Cref{fig:flame-schematic} provides a schematic of the simulation configuration.
		Both DNS and LES are in a rectangular spatial domain:
		\begin{equation}
			\bf x = [x_1, x_2, x_3]^\top \in [-L/2,L/2] \times [-L/2,L/2] \times [-8L,8L]
		\end{equation}
		with $L=1$, uniformly discretized as $\Delta x_1 = \Delta x_2 = \Delta x_3$.
		The flame will be statistically planar on the $x_1 x_2$-plane and the normal to the $x_3$-direction.
		The grid is staggered where $\bf u$ and $\rho \bf u$ are stored at cell faces normal to the coordinate axes, and the other state variables are stored at cell centers~\cite{harlow1965numerical,nagarajan2003robust,boersma2005staggered}.
		Second-order finite differences are used to compute the spatial derivatives, and second-order linear interpolation is used to interpolate from a cell center to cell faces or vice versa.
		In the total energy equation and the species equation, scalar advection is upwind to suppress spurious oscillations, where scalar values ($T$ and $Y$) are reconstructed on the cell face.
		Following Larrouturou~\cite{larrouturou1991preserve}, this is done by piecewise linear reconstruction with Superbee~\cite{roe1985some} limiting.
		Momenta $\rho \bf u$ need not be reconstructed as they already exist on cell faces.

		The turbulent flame simulation is developed in stages: (1) a non-reacting forced isotropic turbulence is generated~\cite{eswaran1988examination} (\cref{sec:forcing}); (2) once the isotropic turbulence is statistically stationary, a laminar flame is superimposed onto the turbulent flow field by re-initializing the $\rho$, $T$, and $Y$ fields; and (3) the turbulent flame is allowed to become statistically stationary before data are used.
		The flame propagates in the $+x_3$-direction so $Y=1$ near the $-x_3$-boundary and $Y=0$ near the $+x_3$-boundary.
		The spatial domain is set as periodic in all dimensions, and the discontinuity at the $x_3$-boundary is relieved in a sponge layer~\cite{freund1997proposed} on each end (details of this are in~\cref{sec:sponge}).
		The turbulence forcing and the sponge are implemented as source terms in the governing equations:
		\begin{equation} \label{eqn:reactingNS+forcing}
			\partial_t \vec q + \nabla \cdot \vec{\bf f}(\vec q) = \vec s(\vec q) + \vec s_\mathrm{forcing} + \vec s_\mathrm{sponge}.
		\end{equation}

		\subsubsection{Turbulence forcing} \label{sec:forcing}
			The turbulence forcing scheme was developed by Eswaran and Pope~\cite{eswaran1988examination} and is formulated as
			\begin{equation}
				\vec s_\mathrm{forcing} = \begin{bmatrix}
					0 \\ \rho \bf b \\ \rho \bf u \cdot \bf b - \rho \mathbb E[\dot e] \\ 0
				\end{bmatrix},
			\end{equation}
			where $\bf b$ is sampled from an Ornstein--Uhlenbeck (OU) process every time step, and $\mathbb E[\dot e]$ is the expected energy injection rate by the forcing.
			Subtracting constant $\mathbb E[\dot e]$ from the total energy keeps the internal energy stationary, so the forcing only excites the kinetic energy.
			It has been shown~\cite{eswaran1988examination} that $\mathbb E[\dot e] = 4 N_{\bm \kappa} \sigma_\scf^2 \tau_\scf$, where $N_{\bm \kappa}$ is the total number of discrete wavenumber vectors $\bm \kappa$ being forced, $\sigma_\scf^2$ the variance of the OU process, and $\tau_\scf$ its correlation time.
			Once the turbulence is statistically stationary, it can be assumed that the kinetic energy dissipation rate converges to $\mathbb E[\dot e]$, meaning that a desired value of dissipation rate can be achieved by setting the energy injection rate.
			As $\tau_\scf$ is the most energetic time scale of isotropic turbulence, it is also an approximate integral time scale, so an approximate integral length scale can defined as $\ell_\scf = (\mathbb E[\dot e] \tau_\scf^3)^{1/2}$.
			Since length scales and velocity scales are nondimensionalized by $\ell_\scf$ and $u_\scf \equiv \ell_\scf / \tau_\scf$ respectively, $\mathbb E[\dot e] = 1$.
			Forcing is applied to $N_{\bm \kappa} = 26$ modes: $\bm \kappa/\kappa_0 \in \{ (\pm 1, 0, 0), (0, \pm 1, 0), (0, 0, \pm 1), (\pm 1, \pm 1, 0), (\pm 1, 0, \pm 1), (0, \pm 1, \pm 1), (\pm 1, \pm 1, \pm 1) \}$, where $\kappa_0 = 2\pi/L$, so $\sigma_\scf^2 = 1/104$.

		\subsubsection{Absorbing sponge zone boundary regions} \label{sec:sponge}
			The sponge zone drives the state toward a target condition with the damping term $\vec s_\mathrm{sponge} = -\zeta(\bf x) (\vec q - \vec q_\infty)$, where the damping strength is
			\begin{equation}
				\zeta(\bf x) = \begin{cases}
					200 [(x_3^- - x_3)/\ell^-]^3 & \phantom{x_3^- <} \ x_3 \leq x_3^- \\
					0 & x_3^- < x_3 < x_3^+ \\
					200 [(x_3 - x_3^+)/\ell^+]^3 & x_3^+ \leq \ x_3
				\end{cases}. \label{eqn:damping-profile}
			\end{equation}
			In~\cref{eqn:damping-profile}, $x_3^\pm$ is the $x_3$-coordinate where the sponge layer starts, and $\ell^\pm$ is the thickness of the sponge layer.
			Since sound propagates faster in the burnt region than in the unburnt region, the sponge layer is set to be thicker in the burnt region: $\ell^+ = 3L$ and $\ell^-=L$, which yields $x_3^+=5L$ and $x_3^-=-7L$.

			The target conditions are
			\begin{equation}
				\vec q_\infty = \begin{cases}
					\vec q_\infty^{\,-} & x_3 \leq x_3^- \\
					\vec q_\infty^{\,+} & x_3 \geq x_3^+
				\end{cases}
			\end{equation}
			where
			\begin{equation}
				\vec q^{\,+}_\infty = \begin{bmatrix}
					\rho_b \\ 0 \\ 0 \\ \rho_b [\alpha s_L/(1-\alpha) + s_T] \\ T_0/\gamma + \rho_b [\alpha s_L/(1-\alpha) + s_T]^2/2 \\ 0
				\end{bmatrix} \quad \textrm{and} \quad \vec q^{\,-}_\infty = \begin{bmatrix}
					\rho_0 \\ 0 \\ 0 \\ \rho_0 s_T \\ T_0/\gamma + \rho_0 s_T^2/2 \\ \rho_0
				\end{bmatrix},
			\end{equation}
			with $\rho_0$ and $\rho_b=(1-\alpha)\rho_0$ the densities of the unburnt and burnt gas respectively, $s_L$ the laminar flame speed which can be estimated~\cite{williams2018combustion}:
			\begin{equation}
				s_L = \sqrt{\frac{2A(1-\alpha)^{2-n}}{\mathrm{Re\,Pr} \beta^2}},
			\end{equation}
			and $s_T$ is the average turbulent flame speed (related to the time-averaged global fuel consumption rate)
			\begin{equation}
				s_T = \frac{1}{\rho_0 L^2 \tau_\mathrm{avg}} \int_{t_0}^{t_0+\tau_\mathrm{avg}} \int_\Omega \dot \omega(\bf x, t) \, d\bf x \, dt.
			\end{equation}
			Time-averaging is over time $\tau_\mathrm{avg}=10$, and $\Omega$ is the physical region of the domain, which is the entire domain aside from the buffer zones.
			Iterations showed that $s_T \approx 2.5$, so setting $s_T = 2.5$ approximately anchors the flame in the middle of the domain.

	\subsection{NN formulation} \label{sec:nnform}
		For any unconstrained NN in~\cref{eqn:nn-flame-const} and~\cref{eqn:nn-flame-unconst}, both the input and the output are stored at cell centers.
		At a grid point $\bf x$, its input is composed of discrete values of primitive variables $\vec q_p = [\rho, \bf u, \hat T, Y]^\top$ at $\bf x$ and the neighboring points:
		\begin{equation}
			\vec{\mathcal I}[\vec q_p](\bf x) = \begin{bmatrix}
				\{\rho(\bf x); \hat T(\bf x); Y(\bf x)\} \in \mathbb R^3 \\
				 \{ \rho(\bf x \pm \bm \Delta_i) - \rho(\bf x) \}_{i=1}^3 \in \mathbb R^6 \\
				 \{ \bf u(\bf x \pm \bm \Delta_i) - \bf u(\bf x) \}_{i=1}^3 \in \mathbb R^{18} \\
				 \{ \hat T(\bf x \pm \bm \Delta_i) - \hat T(\bf x) \}_{i=1}^3 \in \mathbb R^6 \\
				 \{ Y(\bf x \pm \bm \Delta_i) - Y(\bf x) \}_{i=1}^3 \in \mathbb R^6
			\end{bmatrix} \in \mathbb R^{39},
		\end{equation}
		where $\bm \Delta_i = \Delta x_i \bf e_i$ with $\bf e_i$ the unit vector in $x_i$.
		The input consists of 3 scalar variables $[\rho, \hat T, Y]^\top$ at $\bf x$ and both forward (between $\bf x + \bm \Delta_i$ and $\bf x$) and backward (between $\bf x - \bm \Delta_i$ and $\bf x$) differences of 6 primitive variables $\vec q_p$ in 3 spatial dimensions $[x_1, x_2, x_3]^\top$, yielding 39 total input variables per mesh point.
		Pointwise values of $\bf u$ are excluded for Galilean invariance.
		Therefore, $f_{\NN,ij}$, $s_\NN$, and $\{ f_{\NN,Ti} \}_{i=1}^3$ and $\{ f_{\NN,Yi} \}_{i=1}^3$ for unconstrained models are all $\in \mathcal N(39 \to 50 \to 1)$.
		As outputs are originally at cell centers, those of $f_{\NN,ij}$ are linearly interpolated to cell faces.

		Constrained models, $\vec f_{\NN,ci}$ and $\vec \nu_{\NN,ci}^{\,\pm}$, are evalauted at cell faces.
		The NN $\vec f_{\NN,ci}$ takes only the reactive scalars $\vec q_c = [\hat T, Y]^\top$ as the input as
		\begin{equation}
			\vec f_{\NN,ci}(\vec q_c^{\,\pm}(\bf x + \bm \Delta_i/2)) \in \mathcal N(2 \to 50 \to 2),
		\end{equation}
		where
		\begin{equation}
			\begin{split}
				\vec q_c^{\,+}(\bf x + \bm \Delta_i/2) &= \vec q_c(\bf x + \bm \Delta_i) - 0.5 \phi[\vec r_i(\bf x + \bm \Delta_i)] [\vec q_c(\bf x + 2 \bm \Delta_i) - \vec q_c(\bf x + \bm \Delta_i)] \quad \mathrm{and} \\
				\vec q_c^{\,-}(\bf x + \bm \Delta_i/2) &= \vec q_c(\bf x) + 0.5 \phi[\vec r_i(\bf x)] [\vec q_c(\bf x + \bm \Delta_i) - \vec q_c(\bf x)]
			\end{split}
		\end{equation}
		are the slope-limited reconstructions with the slope ratio
		\begin{equation}
			\vec r_i (\bf x) = \frac{\vec q_c(\bf x) - \vec q_c(\bf x - \bm \Delta_i)}{\vec q_c(\bf x + \bm \Delta_i) - \vec q_c(\bf x)}.
		\end{equation}
		For the NN diffusivity, $\vec \nu_{\NN,ci}^{\,\pm}$ takes the 39 inputs
		\begin{equation}
			\vec \nu_{\NN,ci}^{\,\pm} (\bf x + \bm \Delta_i/2) = \vec \nu_{\NN,ci}^{\,\pm} [\vec{\mathcal I}[\vec q_p](\bf x + \bm \Delta_i/2)] \in \mathcal N(39 \to 50 \to 2),
		\end{equation}
		where $\vec{\mathcal I}[\vec q_p](\bf x + \bm \Delta_i/2) = [\vec{\mathcal I}[\vec q_p](\bf x) + \vec{\mathcal I}[\vec q_p](\bf x + \bm \Delta_i)]/2$.

	\subsection{Adjoint-based stochastic gradient descent}	\label{sec:adjoint}
		In the one-dimensional demonstrations, the end-to-end sensitivity $\mathrm d \mathcal J / \mathrm d \vec \theta$ was computed using the full AD supported by PyTorch~\cite{paszke2019pytorch}.
		However, computing sensitivities solely based on AD can be slow and require excessive memory for a complex system, particularly for $\textit{a posteriori}$ NN training for three-dimensional simulations.
		An efficient alternative is to directly formulate and solve adjoint governing equations~\cite{sirignano2020dpm,macart2021embedded}.
		This is identical to the full AD method in that they both solve the dual of the primal problem~\cref{eqn:reactingNS+forcing}.
		However, direct solving of adjoint equations is less expensive than the full AD since it leverages the prior knowledge on all the operations, whereas the full AD method tracks all the operations \textit{ad hoc} with unnecessary intermediate values or gradient-tracking features, adding cost.

		For a PDE solution $\vec q(\bf x, t; \vec \theta\,)$ on $t \in [t_0^{(m)}, t_0^{(m)} + t_f]$ starting from a time stamp $t_0^{(m)}$ as in~\cref{eqn:loss-flame}, sensitivities are computed using adjoints as
		\begin{equation}
			\dv{\mathcal J^{(m)}}{\vec \theta} = \int_\Omega \left( \sum_{n=0}^{N_t-1} \sum_{s=1}^4 w_s \vec q^{\,\dagger n,s} \cdot \pdv{\vec R}{\vec \theta} \eval^{n,s-1} \right) \, d \bf x,
		\end{equation}
		where $\vec R$ is the RHS of~\cref{eqn:reactingNS+forcing} including the NN terms.
		The adjoint $\vec q^{\,\dagger}$ of $\vec q$ is obtained by solving the adjoint governing equations backward in time,
		\begin{equation}
			\begin{split}
				\vec q^{\,\dagger n,3} &= \vec q^{\,\dagger n,4} + \frac{\Delta t}{2} \vec q^{\,\dagger n,4} \cdot \pdv{\vec R}{\vec q} \eval^{n,3} \\
				\vec q^{\,\dagger n,2} &= \vec q^{\,\dagger n,4} + \frac{\Delta t}{2} \vec q^{\,\dagger n,3} \cdot \pdv{\vec R}{\vec q} \eval^{n,2} \\
				\vec q^{\,\dagger n,1} &= \vec q^{\,\dagger n,4} + \Delta t \vec q^{\,\dagger n,2} \cdot \pdv{\vec R}{\vec q} \eval^{n,1} \\
				\vec q^{\,\dagger n-1,4} = \vec q^{\,\dagger n,0} &= \vec q^{\,\dagger n, 4} + \Delta t \sum_{s=1}^4 w_s \vec q^{\,\dagger n,s} \cdot \pdv{\vec R}{\vec q} \eval^{n,s-1},
			\end{split}
		\end{equation}
		starting from the final condition
		\begin{equation}
			\vec q^{\,\dagger N_t-1,4} = \vec q^{\,\dagger N_t,0} = \pdv{D^{(m)}(\bf x)}{\vec q^{N_t,0}}
		\end{equation}
		at time step $N_t$, which corresponds to $t = t_0^{(m)} + t_f$, until $n=0$ at $t = t_0^{(m)}$.
		The pointwise mismatch $D^{(m)}(\bf x) = \Vert \vec q_p(\bf x, t_0^{(m)} + t_f; \vec \theta\,) - \vec q_{p,e} (\bf x, t_0^{(m)} + t_f) \Vert_2^2$ contributes to the loss as
		\begin{equation}
			\mathcal J^{(m)} = \int_\Omega D^{(m)}(\bf x) \, d \bf x.
		\end{equation}
		The Jacobian $\partial \vec R/\partial \vec q$ and the $\vec q^{\,\dagger N_t,0} = \partial D^{(m)}/\partial \vec q$ are evaluated based on the fully discretized form of~\cref{eqn:reactingNS+forcing}, which makes the computed sensitivity exact for the discrete system~\cite{vishnampet2015practical}.

\section{Design Choices for Machine Learning} \label{sec:appendix2}
\subsection{Architecture}
The architecture of the model~\cref{eqn:nn-architecture} comes from Sirignano and Spiliopoulos~\cite{sirignano2018dgm} for sequential PDE solutions, mimicking the long short-term memory (LSTM) models.
In addition to hidden layers and nonlinear activation functions, which are essential components of deep NN models, it has an elementwise product between hidden layers and a nonlinear map of the input.
This feature is thought to be motivated by the LSTM architecture and brings additional extent of nonlinearity to the model.
Even in the case where the tanh activation functions are saturated, the elementwise products guarantee the model to be capable of learning at least third-degree polynomials.

For most of our demonstrations from~\cref{sec:advection} to~\cref{sec:antidiffusion}, the true flux functions $f_e$ can be exactly expressed with third-degree polynomials:
\begin{gather}
	f_e(q) = q \quad \text{(one-dimensional advection)} \quad \text{and} \\
	f_e(q) = q^2/2 \quad \text{(one-dimensional Burgers)}.
\end{gather}
Therefore, we expect the architecture~\cref{eqn:nn-architecture} to be sufficiently complex to learn such physics.
The true flux function for the one-dimensional Euler equations may not be expressed with only the third-degree polynomials as
\begin{equation}
\vec f_e(\vec q) = \begin{bmatrix}
	\rho u \\ \rho u^2 + p \\ u(E + p)
\end{bmatrix} = \begin{bmatrix}
	q_1 \\ (3-\gamma) q_1^2 / (2q_0) + (\gamma-1) q_2 \\ \gamma q_1 q_2 / q_0 - (\gamma-1) q_1^3 / (2 q_0^2)
\end{bmatrix} \quad \text{for} \quad \vec q = \begin{bmatrix}
q_0 & q_1 & q_2
\end{bmatrix}^\top
\end{equation}
is up to fifth-degree, including reciprocals.
Nonetheless, with the nonlinear activation functions adding more nonlinearities, our model is deemed to be capable of learning all the nonlinear fluxes that we considered.

Of course, it is unknown to what extent it approximates an unknown flux, as in~\cref{sec:premixed}.
However, we note that no model is guaranteed to be perfect, and the rationale behind most ML methods is to make the model sufficiently complex to maximize its approximating power in the spirit of the universal approximation theorem~\cite{cybenko1989approximation}.

\subsection{Hyperparameters}
Although numerous studies exist that look for optimal set of hyperparameters, a common workflow of ML modeling approach is to first start with a working model, then to prune it.
For all demonstrations in this paper, it might be worthwhile to prune them, but the goal of this paper is to demonstrate limiting.
Hence, the number of layers and units, and other hyperparameters, including the learning rate and initial values, are selected by trial and error.
Since the loss function for deep NN models is non-convex and there is generally no optimal choice of scheduling the learning rate, common choices are the adaptive learning rate with under-relaxation, such as RMSprop~\cite{tieleman2012lecture} and Adam~\cite{kingma2014adam}, based on rule of thumb.
We have tested different types of optimizers for each demonstration cases, and we have selected the one that was the most effective, though we did not included results from other optimizers since they are not essential to our demonstrations and conclusions.
\bibliographystyle{ieeetr}

\end{document}